\documentclass[10pt,journal,compsoc]{IEEEtran}
\usepackage[utf8]{inputenc} 
\usepackage[T1]{fontenc}    
\usepackage{hyperref}       
\usepackage{url}            
\usepackage{graphicx}

\usepackage{amsmath,amsthm,amssymb,amsfonts}
\usepackage{color}

\usepackage{url}
\usepackage{array}
\newcommand{\etal}{\emph{et al.~}}
\newcommand{\etalnospace}{\emph{et al.}}
\DeclareMathOperator*{\argmin}{arg\,min}
\newcommand{\fig}[2]{\includegraphics[width=#2cm]{images/#1}}
\newcommand{\parbasic}[1]{\noindent\textbf{#1}}
\newcommand{\hn}{\hspace{-0.3cm}}
\newcommand{\hnb}{\hspace{-0.15cm}}

\newcommand{\revised}[1]{#1}
\newcommand{\revisedcont}[1]{#1}
\newcommand{\revisedtag}[2]{#2}
\newcommand{\captiontag}[1]{}
\newcommand{\revisedremove}[1]{}
\begin{document}

\title{Detecting Deficient Coverage in Colonoscopies}
\author{Daniel Freedman, Yochai Blau, Liran Katzir, Amit Aides, Ilan Shimshoni, Danny Veikherman, Tomer Golany, Ariel Gordon, Greg Corrado, Yossi Matias, and Ehud Rivlin
\thanks{Daniel Freedman, Yochai Blau, Liran Katzir, Amit Aides, Ilan Shimshoni, Danny Veikherman, Tomer Golany, Yossi Matias, and Ehud Rivlin are with Google Research, Israel (e-mail: danielfreedman@google.com).  Ariel Gordon and Greg Corrado are with Google Research, USA.}}

\IEEEtitleabstractindextext{%
\begin{abstract}
Colonoscopy is the tool of choice for preventing Colorectal Cancer, by detecting and removing polyps before they become cancerous.  However, colonoscopy is hampered by the fact that endoscopists routinely miss 22-28\% of polyps.  While some of these missed polyps appear in the endoscopist's field of view, others are missed simply because of substandard coverage of the procedure, i.e. not all of the colon is seen.  This paper attempts to rectify the problem of substandard coverage in colonoscopy through the introduction of the C2D2 (\underline{C}olonoscopy \underline{C}overage \underline{D}eficiency via \underline{D}epth) algorithm which detects deficient coverage, and can thereby alert the endoscopist to revisit a given area. More specifically, C2D2 consists of two separate algorithms: the first performs depth estimation of the colon given an ordinary RGB video stream; while the second computes coverage given these depth estimates.  Rather than compute coverage for the entire colon, our algorithm computes coverage locally, on a segment-by-segment basis; C2D2 can then indicate in real-time whether a particular area of the colon has suffered from deficient coverage, and if so the endoscopist can return to that area.  Our coverage algorithm is the first such algorithm to be evaluated in a large-scale way; while our depth estimation technique is the first calibration-free unsupervised method applied to colonoscopies.
\revisedtag{1}{
The C2D2 algorithm achieves state of the art results in the detection of deficient coverage.  On synthetic sequences with ground truth, it is 2.4 times more accurate than human experts; while on real sequences, C2D2 achieves a 93.0\% agreement with experts.
}
\end{abstract}

}

\maketitle
\IEEEdisplaynontitleabstractindextext
\IEEEpeerreviewmaketitle

\IEEEraisesectionheading{\section{Introduction}\label{sec:introduction}}
\IEEEPARstart{C}{olorectal} Cancer (CRC) is a global health problem, resulting in an estimated 900K deaths per year \cite{colorectal2018fact}; it is the second deadliest cancer in the United States \cite{cancer2019cancer}.  CRC is different from other leading cancers in that it is preventable.  Specifically, polyps, which are small precancerous dwellings in the colon, may be detected and removed before they actually become cancerous.  Colonoscopy is considered the gold standard procedure for the detection and removal of polyps.  Whereas fecal immunochemical and related tests may detect CRC once it has become malignant, colonoscopy is able to detect the polyps in their precancerous stage, thereby preventing cancer from developing.  And in contrast to wireless capsule endoscopy, colonoscopy can not only detect, but also remove, polyps.

Unfortunately, the literature indicates that endoscopists miss on average 22-28\% of polyps during colonoscopies, which includes 20-24\% of adenomas \cite{leufkens2012factors}.  (An adenoma is a polyp which has the potential to become cancerous; this is in contrast to a hyperplastic polyp, which is benign.)  There is therefore room for improvement in polyp detection during colonoscopies.  The importance of these missed polyps can be quantified in terms of the rate of interval CRC, defined as a CRC that is diagnosed within 60 months of a negative colonoscopy \cite{lee2017clinical}. In particular, it is estimated that a 1\% increase in the Adenoma Detection Rate (ADR, defined as the fraction of procedures in which a physician discovers at least one polyp) can lead to a 6\% decrease in the rate of interval CRC \cite{kaminski2017increased}.

\begin{figure}[t]
  \centering
  \fig{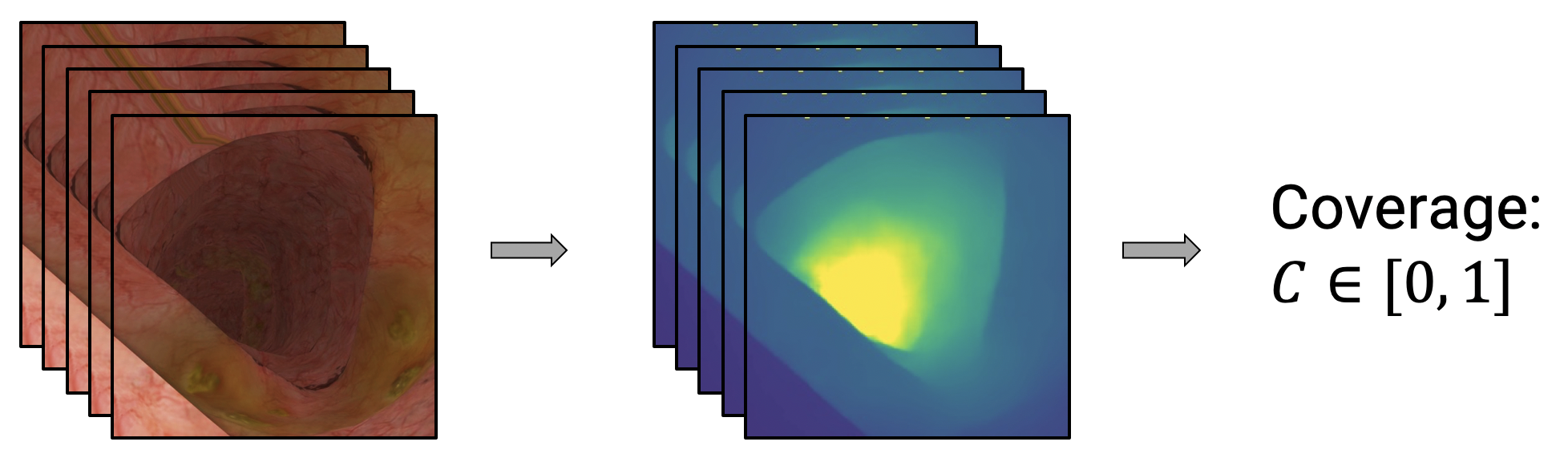}{8.5}
  \caption{
  \captiontag{2}
  \revised{
  Our algorithm computes a depth image from a given RGB image.  Then, based on the computation of a depth image sequence from a video sequence, the algorithm can compute local coverage, and therefore detect where the coverage has been deficient and a second look is required.
  }
  }
  \label{fig:teaser}
\end{figure}

It is therefore imperative to decrease the polyp miss-rate during colonoscopies.  There are several factors which lead endoscopists to miss polyps.  Some factors, such as bowel preparation, can only be addressed by better patient compliance with the preparatory process.  But other factors, such as endoscopist fatigue and endoscopist experience level, can be aided by AI-based real-time decision support systems.  In particular, given a well-prepped bowel, there are two principal reasons why an endoscopist might miss a polyp: (1) the polyp appears in the field of view, but the endoscopist misses it, perhaps due to the difficulty of detection, e.g. the polyp may be very small or flat; (2) the polyp does not appear in the field of view, as the endoscopist has not properly covered the relevant area during the procedure.  Note that these two reasons are orthogonal, and demand different types of computer vision-based solutions.  In terms of the first reason, polyp detection systems such as \cite{urban2018deep,wang2019real} have been shown to be quite effective.  In this paper we choose to focus on the second reason for missing polyps: deficient coverage.

As we have explained, our main motivation for computing coverage is to detect when said coverage is deficient, and thereby decrease the polyp miss-rate.  There is a secondary motivation, however, which is that coverage is a performance measure, similar to ADR, by which endoscopists can be graded.  The consensus within the field of gastroenterology is that for a procedure to be effective, 90-95\% of the colon ought to have been covered \cite{rex2007best}.  An algorithm for computing coverage could therefore be used both for alerting the endoscopist to missed regions, as well as for measuring the endoscopist's performance.

We refer to our approach to coverage computation as the C2D2 algorithm: \underline{C}olonoscopy \underline{C}overage \underline{D}eficiency via \underline{D}epth.  C2D2 consists of two separate algorithms.  The first performs depth estimation of the colon given an ordinary RGB video stream; while the second computes coverage given these depth estimates.  We now outline each of these in turn.  The method we use for depth estimation is based on a deep learning approach, in which the network maps RGB images directly to depth images.  One advantage presented by any network-based solution is that it allows for the depth estimation algorithm to run in real-time.  However, the particular deep learning approach we use offers two further benefits.  First, the approach relies only on unsupervised data; thus, one can learn directly from colonoscopy videos without the need for any supervisory signal.  Alternative techniques are often based on learning from synthetic data, for which there is depth-supervision, e.g. \cite{wang2017deepvo,turan2018deep}; however, this entails the need for domain adaptation, which we avoid.  Second, our method is calibration-free: it learns the camera intrinsics as part of the algorithm.  This is particularly important, as acquiring the intrinsic parameters of a given endoscope is not straightforward; and each colonoscopy will use a different endoscope, entailing different parameters.

Given the depth estimates, C2D2 can then compute coverage, and detect when it is deficient.
\revisedtag{3}{
Coverage is computed on a segment-by-segment basis; we will make the definition of coverage precise in Section \ref{sec:coverage}, but for now, it may be thought of as a scalar in $[0, 1]$ which measures what fraction of the colon has been viewed in any given segment.
}
The coverage algorithm is also based on deep learning, but due to the particular character of the problem -- that is, the impossibility of ground truth labelling on real colonoscopy sequences -- we must train on synthetic sequences.  However, in the final analysis the coverage algorithm must also work on real sequences.  The joint requirements of training on synthetic sequences but inference on real sequences leads to a novel two network architecture, with a corresponding two stage training process.

To the best of our knowledge, the C2D2 algorithm is the first to be evaluated on a large scale test set  (previous work has tended to perform evaluation on a handful of examples e.g. \cite{ma2019real}).  We provide quantitative performance results on a collection of 561 synthetic sequences with ground truth.  Our results show that on this set, C2D2 outperforms physicians by a factor of $2.4$, according to the Mean Average Error (MAE) of coverage.  
\revisedremove{
A direct implication of these results is that physicians are not particularly accurate in estimating coverage (they have a high MAE); thus, one cannot use physicians to provide ground truth labels for real sequences.  As a result, it is not possible to quantitatively assess performance on real sequences.  Nevertheless, we can provide \emph{qualitative} performance results on real sequences; while ground truth is not available for such sequences, we show that C2D2 outputs highly plausible coverage scores that agree with the eyeball test.
}
\revisedtag{4}{
On real sequences, no ground truth is available.  Instead, we show that on a set of 301 real sequences, C2D2 achieves a 93.0\% agreement with human experts.  We also provide qualitative performance results on real sequences, which show that C2D2 outputs highly plausible coverage scores that agree with the eyeball test.
}

These results demonstrate the value of the C2D2 system: the computation of coverage in general, and detection of deficient coverage in particular, are highly geometric tasks.  In such tasks, it is often the case that computers outperform humans, and this is borne out by our results.  In many AI tasks, the goal is simply to do as well as human experts; in our case, the system outperforms humans, and this is where its true value lies.

To summarize, our contributions are fourfold:
\begin{enumerate}
    \item We propose a novel approach to coverage, which is implemented using a two network architecture with a corresponding two stage training process.
    \item We present the first calibration-free unsupervised method for depth estimation applied to colonoscopies.
    \item The combined C2D2 system is the first coverage system to be evaluated in a large-scale way, and outperforms human experts by a wide margin on coverage tasks.
    \item \revisedtag{5}{We release a dataset of synthetic colonoscopy videos on which C2D2 was trained and evaluated.}
\end{enumerate}

The remainder of the paper is organized as follows.  Section \ref{sec:related_work} reviews related work, focusing on coverage and depth estimation within endoscopic procedures; as well as more general modern techniques for SLAM and visual odometry.  Section \ref{sec:depth} presents our technique for calibration-free, unsupervised learning of depth estimation.  Section \ref{sec:coverage} defines precisely the coverage problem we would like to solve, and describes our algorithm for tackling this problems.  Section \ref{sec:results} presents results for both depth estimation as well as coverage, including a detailed description of the new coverage dataset we have collected.  Section \ref{sec:conclusions} concludes.

\section{Related Work}
\label{sec:related_work}
We begin by reviewing the three papers which are, each in a given aspect, most related to the current work.  In \cite{wu2019randomised} and its follow-up paper \cite{chen2019comparing}, Wu \etal propose a blind-spot detector for the EGD (esophagogastroduodenoscopy) procedure, which is an endoscopic procedure focusing on the upper GI tract, including the pharynx, esophagus, stomach, and duodenum.  The idea is, in some sense, to provide a notion of coverage of the upper GI tract; the goal is therefore similar to the goal of the current paper.  Wu \etal divide the upper GI tract into 26 areas, and devise a CNN-based per-frame detector to classify a given image according to which of the 26 regions it belongs to.  A technique based on reinforcement learning is built on top of this classifier in order to encourage temporal consistency.  Wu \etal then verify the usefulness of the real-time system in a randomized controlled trial, and show that endoscopists using the system experience far fewer blind-spots (i.e. regions which are not viewed) than those not using the system, 5.86\% vs. 22.46\%.  The main difference between the approaches of Wu \etal and the current work concerns the notion of coverage which is proposed: in Wu \etalnospace, coverage is in terms of semantic regions, whereas our approach has a much more \emph{geometric} notion of coverage.  We argue that the colon does not have as many varied or differentiated areas as the upper GI tract: the colon, which is quite long, is generally divided into 6 different regions -- the cecum, the ascending colon, the transverse colon, the descending colon, the sigmoid colon, and the rectum.  Therefore a semantic approach to coverage would not work nearly as well in the case of the colon, as the regions are simply too large.  A geometric approach, by contrast, allows for an area of any given size (even a single frame) to be analyzed in terms of coverage, and is therefore a considerably more flexible approach.

A second related work is that of Ma \etal \cite{ma2019real}, which focuses on the problem of depth reconstruction in the colon.  The reconstruction pipeline proposed by \cite{ma2019real} is complex, but essentially consists of two pieces: a deep network piece, which computes both the depth image for the current frame as well as the camera pose; and a more traditional set of SLAM-based geometric procedures for refining the depth and pose, and for stitching the depth images together to create a 3D point cloud.  It is important to note that the network is trained in a supervised fashion; as colonoscopy videos do not come with depth ground truth, a proxy for the ground truth is computed using a separate (non-deep) Structure from Motion algorithm \cite{schonberger2016structure}.  They then use the 3D reconstruction to provide a measure of coverage.  The distinctions between the approach of Ma \etal and our approach are twofold.  First, the depth pipeline of Ma \etal requires supervised data, whereas our technique is purely unsupervised.  Supervision based on Structure from Motion is an interesting idea, but it is difficult to know how effective it is, given that the evaluation in the paper also assumes that the Structure from Motion is the ground truth; it may therefore be that the pipeline has simply learned to compute Structure from Motion depth estimates, rather than true depth.  Second, and more importantly, the coverage algorithm proposed in the paper is not thoroughly evaluated.  Rather, missing region fractions are simply quoted for four colon segments, with the numbers verified by a colonoscopist.  In the current work, we provide a full-scale evaluation of the proposed coverage algorithm.

A final piece of closely related work is the paper of Turan \etal \cite{turan2018unsupervised}, which takes an unsupervised approach to visual odometry and depth reconstruction in the colon, with the primary application being robotic endoscopic capsules.  The approach is based on one of the early unsupervised deep learning techniques for visual odometry and depth estimation for the general computer vision audience \cite{zhou2017unsupervised}.  The differences between our work and this work are fourfold.  First and most importantly, our work does not require known camera intrinsics to work, that is, it is calibration-free.  This is a major difference, as each endoscope has its own intrinsics which are generally not simple to compute.  Second, our work is based on a very recent technique for unsupervised depth estimation \cite{gordon2019depth}, which has been shown to have superior accuracy to \cite{zhou2017unsupervised} on the KITTI dataset.  Third, Turan \etal do not provide any evaluation of their depth estimation algorithm, instead simply showing a few images (they focus instead on evaluating the odometry part of the algorithm).  Fourth, Turan \etal do not discuss issues of coverage.

Other related work has focused on somewhat different versions of the depth reconstruction problem.  Turan \etal \cite{turan2018deep} make use of a supervised deep learning pipeline quite similar to that introduced in \cite{wang2017deepvo} for performing visual odometry and depth reconstruction.  They are most interested in visual odometry, which they evaluate on a dataset they have collected for motion within a porcine stomach.  Both Chen \etal \cite{chen2019slam} and Rau \etal \cite{rau2019implicit} use supervised approaches to depth estimation, where the supervision comes from a synthetic dataset; both use adversarial techniques to ensure that the predicted depth resembles true depth images.  In the case of Chen \etal \cite{chen2019slam}, they further perform stitching on the depth images to yield a single unified point cloud, using the ElasticFusion technique \cite{whelan2016elasticfusion}.  Widya \etal \cite{widya2019whole} perform 3D reconstruction on the whole stomach.  Their technique is based on extraction of SIFT features, followed by a classical Structure from Motion approach \cite{schonberger2016structure,wu2013towards}.  SIFT features are generally known to be problematic in medical images, but the technique works here due to the use of chromoendoscopy, in which the stomach itself is dyed using indigo carmine.  Nevertheless, the reliance on dyeing severely limits the applicability, as chromoendoscopy is not very common.  Slightly older work includes \cite{armin2016automated}, which computes a two-dimensional visibility of the colon; \cite{zhao2016endoscopogram} which is an offline (non-real time) technique that uses classical techniques based on Shape from Shading and Shape from Motion to produce a dense 3D reconstruction; and \cite{hong20143d}, which bases its colon surface reconstruction on the geometry of the Haustral ridges.  Finally, we mention a trio of works \cite{pinheiro2018deep,wang2019novel,aghanouri2019new} whose purpose is to compute pure visual odometry (i.e. pose) without regard to either depth estimation or coverage.

We conclude by briefly surveying the literature related to recent techniques in SLAM, visual odometry, and depth reconstruction intended for the broader computer vision audience.  Earlier work, such as \cite{engel2014lsd,mur2015orb,wang2017stereo}, was geometric in character, and did not use any sort of learning pipeline.  Initial work which applied deep learning techniques, including \cite{wang2017deepvo,almalioglu2019ganvo,yang2018deep,zhou2018deeptam}, did so using the supervision available in such datasets as KITTI.  More recent work \cite{zhou2017unsupervised,li2018undeepvo,yin2018geonet,zhan2018unsupervised,gordon2019depth} has moved to unsupervised deep learning approaches, and is therefore broadly applicable wherever one has video sequences; no depth images are required.  We will go into greater detail regarding one of the most recent (and most successful) of these unsupervised techniques \cite{gordon2019depth} in Section \ref{sec:depth}.

\section{Calibration-Free Unsupervised Depth Estimation}
\label{sec:depth}
We are interested in learning how to estimate a sequence of depth images directly from the corresponding sequence of RGB images of the colonoscopy procedure.  We would like to take a deep learning approach to this problem.  The standard way of tackling this problem requires supervision, in the form of a depth image for each RGB image; for example, such data exists in the case of the KITTI dataset \cite{geiger2013vision}, and is sometimes acquired by equipping the capture device with a depth sensor in addition to a regular camera.  \revisedremove{Unfortunately, in the case of colonoscopy, such datasets do not exist, nor are they likely to exist given the requirement of outfitting the endoscope with an additional depth sensor.}
\revisedtag{6}{
In the case of endoscopy, several such datasets exist, including those used in \cite{mahmoud2018live} based on both CT and dense stereo; those used in  \cite{mahmood2018unsupervised} based on CT; and those used in \cite{maier2014comparative}, based on stereo, structured illumination, and time of flight.
}

In this study, instead, we turn to a purely unsupervised approach to depth estimation.  Over the last two years, a series of papers on deep learning of unsupervised depth estimation have appeared in the computer vision literature \cite{zhou2017unsupervised,garg2016unsupervised,ummenhofer2017demon,mahjourian2018unsupervised,yin2018geonet,gordon2019depth}.  All of these papers are based on essentially the same principle, and differ in their details.  The principle is the \emph{view synthesis loss}, which we now explain.

\begin{figure}[t]
  \centering
  \fbox{\fig{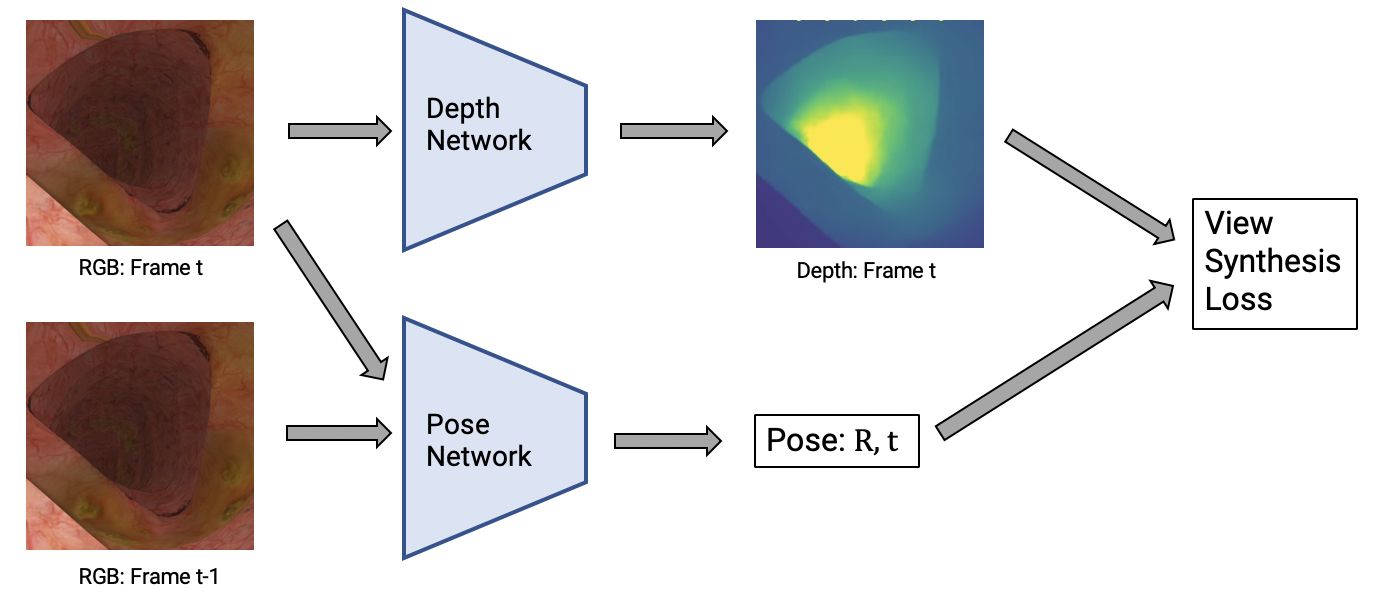}{8.5}}
  \caption{The view synthesis loss and corresponding network architecture.  See accompanying description in the text.}
  \label{fig:view_synthesis}
\end{figure}

\parbasic{General Algorithmic Approach} We proceed as follows.  Instead of trying to solve the problem of unsupervised depth estimation, which seems to be hard, we try to solve an even harder problem: we try to estimate both the depth image and the pose of the camera (sometimes called the visual odometry) simultaneously.  Solving a harder problem seems to be counter-intuitive, but it will afford us an extra benefit in that we can tie the depth and the pose together.  Specifically, we define the pose as the rigid transformation (rotation and translation) from the current frame $t$ to the previous frame $t-1$.  In particular, we imagine that we have two separate networks that we learn: the depth network takes as input the current RGB frame, and outputs the corresponding depth image; while the pose network takes as input the current and previous RGB frames, and outputs the pose.  See Figure \ref{fig:view_synthesis}.  Given this setup, we imagine the following series of steps:
\begin{itemize}
    \item We take the current RGB image $I_t$, and pass it through the depth network to get the current depth image $D_t$.
    \item We take the current and previous RGB frames $I_t$ and $I_{t-1}$, and pass them through the pose network to get the pose, expressed as a rotation matrix $R$ and translation vector $t$.
    \item Considering the depth image $D_t$ as a point cloud, we transform each of the points into the previous frame, according to the standard formula:
      \begin{equation}\label{warp_expand}
         z'p' = KRK^{-1}zp +  K t,
      \end{equation}
    In the above, $p$ and $p'$ are the original and transformed homogeneous coordinates of the pixel, respectively; $z$ and $z'$ are the original and transformed depth of the pixel, respectively; and $K$ is the intrinsic camera matrix:
      \begin{equation}\label{K}
         K =
         \begin{pmatrix}  
           f_x & 0 & x_0 \\ 0 & f_y & y_0 \\ 0 & 0 & 1 
         \end{pmatrix}.
      \end{equation}
    \item Given the projected 3D points, one can then re-render the points using the original RGB values at $I_t$, to get a projected RGB image in the $t-1$ frame, which we label $\hat{I}_{t-1}$.
    \item If the depth and pose have been computed correctly, the original RGB image at $t-1$ and the new projected image $\hat{I}_{t-1}$ ought to be equal!  Thus, our loss is given by $\delta(I_{t-1}, \hat{I}_{t-1})$, where $\delta$ is some metric between images, e.g. $L_1$.  This is the view synthesis loss.
\end{itemize}

\parbasic{Elimination of the Need for Calibration} The view synthesis loss and the corresponding network architecture is illustrated in Figure \ref{fig:view_synthesis}.  One issue, as can be seen by examining Equations (\ref{warp_expand}) and (\ref{K}), is that the camera needs to be calibrated prior to using this technique.  This poses a problem in our case, as each endoscope model (of which there are many) has its own set of intrinsics, and we cannot rely on the possibility of endoscope calibration.  This problem is inherent in most of the work on unsupervised monocular depth estimation, e.g. \cite{zhou2017unsupervised,li2018undeepvo,yin2018geonet,zhan2018unsupervised}.  However, following the very recent technique \cite{gordon2019depth}, we can predict not only depth and pose but also the camera intrinsics ($K$) as well.  This necessitates relatively minor changes to the network architecture: in addition to the depth and pose subnets, there is also a camera intrinsics subnet.  Despite the relative simplicity of implementation, this change is crucial in practice, allowing us to deploy the depth estimation pipeline on any endoscope.

\parbasic{Details and Caveats} Regarding the image metric $\delta$ between $\hat{I}_{t-1}$ and $I_{t-1}$, we use two separate metrics: the $L_1$ difference and the structural similarity (SSIM). In addition to RGB consistency, depth consistency is enforced through an L1 penalty on the difference between the warped depth at the source pixel ($z'$) and the native depth at the target frame. We use the same mechanism as in \cite{gordon2019depth} to avoid enforcing consistency in areas that become occluded or disoccluded upon transitioning between the two frames.

There are two important caveats.  First, in this work we assume that motion is caused primarily due to camera motion.  Under this assumption, $t$ and $R$ are the same for all the pixels in the entire frame, and Equation (\ref{warp_expand}) maps every pixel in a source frame to a target location on the target frame.  In doing so, we are neglecting the non-rigid deformations of the colon.  However, if the non-rigid deformations are sufficiently small between any two frames, this is a reasonable approximation.  Given that the video is taken at 30 fps, meaning that only 33 ms separates two frames, this approximation may hold in practice.  In any case, the results seems to bear out the use of this simplified model.

\revisedtag{7}{
The second caveat is that the depth image is correct only up to a scale factor, i.e. a single scale factor for the entire image.  In principle, the scale factor that is effectively returned by the algorithm should be arbitrary, but in practice, this factor seems to be fairly consistent across long video segments.  From the point of view of coverage, the more critical point is that whatever degree of arbitrariness remains in the scale factor, the coverage algorithm learns to deal with effectively.
}

\section{The C2D2 Algorithm: Computing Coverage}
\label{sec:coverage}
When considering colon coverage, the natural goal is to estimate the fractions of covered and non-covered regions of a complete procedure. Such a formulation of the problem is useful for the physician in terms of a retrospective analysis of a given procedure, as well as general guidance for future procedures.  A more interesting goal, however, is the real time estimation of coverage fraction, on a segment by segment basis; that is, while traversing the colon, the goal is to estimate what fraction of the \emph{current} segment has been covered.  The implications of such a functionality are clear: during the procedure itself, the physician may be alerted to segments with deficient coverage, and can immediately return to review these areas.  This in turn ensures that a higher proportion of polyps will be seen.

We begin this section with a formulation of the coverage problem, including the precise definition of what is meant by segment coverage.  We then discuss our overall approach to the problem, which is based on a two-stage training procedure using synthetic data.  We then discuss each of the stages in turn.  The first stage is a per-frame computation, in which visibility is computed for a given frame.  The second stage takes the network learned in the first stage, and uses it in order to learn per-segment coverage, which is our ultimate goal.

\subsection{Formulation of the Problem}

We begin by defining the coverage in the colon in a mathematically consistent fashion.  A 3D model of a colon consists of the pair $(\mathcal{M}, s)$ where:
\begin{itemize}
    \item $\mathcal{M}$ is a 3D mesh forming the surface of the colon.  
    \item $s(\cdot)$ is a 3D curve, $s:[0,L] \to \mathbb{R}^3$, traversing the whole colon, and lying in the center of the colon. The curve is parameterized by its distance $\ell$ along the curve from the beginning of the colon (rectum). This curve is known as the \emph{lumen} of the colon.
\end{itemize}
We can associate to each point $m$ on the mesh $\mathcal{M}$ the closest point to it on the lumen and its corresponding parameter value:
\[
\ell^*(m) = \argmin_{\ell \in [0, L]} \| m - s(\ell) \|
\]
Similarly, a given camera position $p$ within the colon can also be associated to its nearest point on the lumen, and for ease of notation we also denote the corresponding parameter value as $\ell^*(p)$.

\begin{figure}[t]
  \centering
  \fig{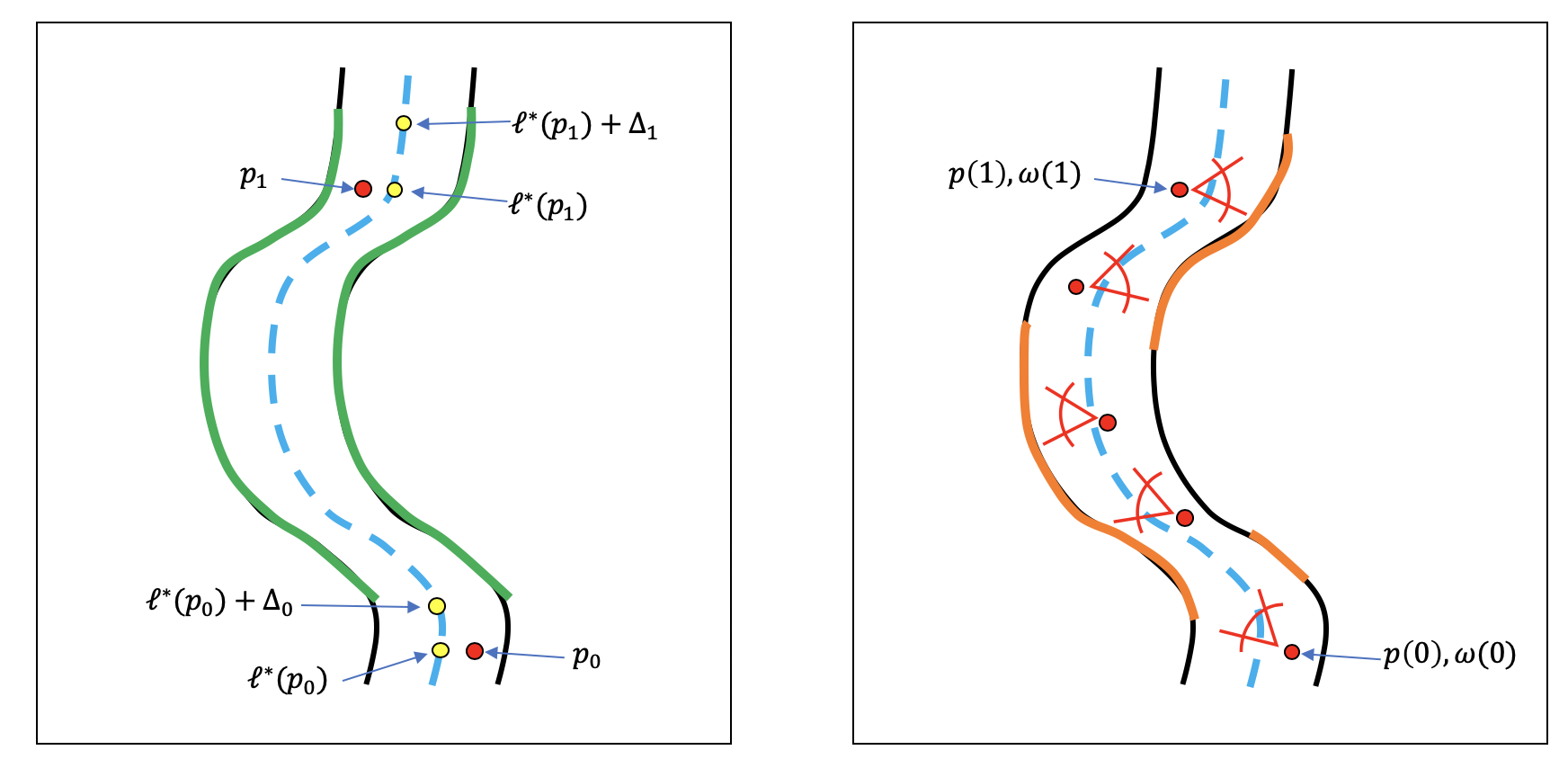}{8.5}
  \caption{\captiontag{9}\revised{Illustration of coverage.  In both figures, the colon is shown in 2D in black, and the lumen is the dashed blue curve; camera locations are shown as red dots.  Several points on the lumen are denoted in yellow, along with their corresponding parameter values $\ell \in [0, L]$.  Left: the maximal set of visible points $\mathcal{V}(p_0, p_1)$ is shown as the green curves, which are a subset of the colon surface (black).  Right: for a given trajectory, illustrated by red camera locations and viewing angles (at a discrete set of points which subsample of the trajectory), the set of actually visible points $\mathcal{A}(p(\cdot), \omega(\cdot))$ is shown in orange.  The coverage is then the ratio of the area of the orange points to the area of the green points.}}
  \label{fig:coverage_diagram}
\end{figure}

Now, consider a segment of a colonoscopy video, where the initial and final camera positions are $p_0$ and $p_1$.  Assuming that the path the camera takes is monotonic -- that is, the camera is moving from the end of the colon (the cecum) towards the rectum pointing in the direction of the cecum -- then the maximal set of points on the colon that can be visible is given by
\begin{equation}
    \mathcal{V}(p_0, p_1) = \{m \in M : \ell^*(p_0) + \Delta_0 \le \ell^*(m) \le \ell^*(p_1) + \Delta_1 \}
    \label{eq:maximally_visible}
\end{equation}
\revisedtag{8}{
In the above, $\Delta_0$ accounts for the viewing angle of the camera: due to the fact that the camera has a field of view that is less than $180^\circ$, one cannot see details that are immediately to the side of the initial camera location $p_0$.  Hence $\Delta_0 > 0$; and as the viewing angle becomes smaller, $\Delta_0$ becomes larger.  By contrast, $\Delta_1$ accounts for the fact that the image taken from the deepest point on the sequence can
}
\revisedcont{
view deeper points on the colon.  Specifically, the deepest points are ones whose closest point on the lumen is a further distance $\Delta_1$ from the closest point on the lumen to the final camera position $p_1$.  These concepts are illustrated in the left subfigure of Figure \ref{fig:coverage_diagram}.
}

\revisedremove{In the above, $\Delta_0$ accounts for the viewing angle of the camera, while $\Delta_1$ accounts for the fact that the image taken from the deepest point on the sequence can view deeper points on the colon.  Note that a special case of the coverage can be computed when only a single frame is considered. In this case $p_0 = p_1$, and we compute $\mathcal{V}(p_0, p_0)$.}

The above computation deals with the maximal set of visible points.  In practice, not all points are viewed, and this is what leads to deficient coverage.  Specifically, given a particular camera position $p \in \mathbb{R}^3$ and orientation $\omega \in \Omega$, we can define the actual set of points on $\mathcal{M}$ that are visible, which we denote $\mathcal{A}(p, \omega)$.  This is computed by rendering the image given the mesh and the camera pose (position and orientation), given the camera's internal calibration parameters (focal length and principal point); one can then verify which points on $\mathcal{M}$ appear in the rendered image, and these are in the points in $\mathcal{A}(p, \omega)$.  Given a full camera trajectory, which we denote by $p:[0,1] \to \mathbb{R}^3$ and $\omega:[0,1] \to \Omega$, the set of actually visible points for the whole trajectory is simply given by
\begin{equation}
    \mathcal{A}(p(\cdot), \omega(\cdot)) = \bigcup_{t \in [0,1]} \mathcal{A}(p(t), \omega(t))
    \label{eq:actually_visible}
\end{equation}
\revisedtag{10}{These concepts are illustrated in the right subfigure of Figure \ref{fig:coverage_diagram}.}

\revisedtag{11}{Finally, given a particular camera trajectory $(p(\cdot), \omega(\cdot))$, the coverage is defined as the ratio of actually visible points to maximally visible points.  That is, combining Equations (\ref{eq:maximally_visible}) and (\ref{eq:actually_visible}), we define the coverage as
\begin{equation}
    \mathcal{C}(p(\cdot), \omega(\cdot)) = \frac{\mu[\mathcal{A}(p(\cdot), \omega(\cdot))]}{\mu[\mathcal{V}(p(0), p(1))]}
    \label{eq:coverage}
\end{equation}
where $\mu$ is the standard measure.  It is important to note that using the standard measure implies that coverage is based on the fraction of surface area, rather than the fraction of pixels.}  Note also that in practice, if the vertices on the mesh are sufficiently dense, then one can simply count the vertices in both $\mathcal{A}(p(\cdot), \omega(\cdot))$ and $\mathcal{V}(p(0), p(1))$.

\revisedtag{12}{
We end this section by noting that it is common practice in colonoscopy screening for the endoscopist to retroflex the endoscope during withdrawal to examine proximal sides of folds and closely examine the mucosa.  Effectively, this means that the endoscopist examines one side of the colon wall, immediately followed by an examination of the other side of the wall.  A natural question might be: how does this affect the definition of coverage in Equation (\ref{eq:coverage})?  The answer is that the definition of coverage can accommodate this situation without difficulty.  The video segment in question contains both sides of the wall, which implies that the set of actually visible points $\mathcal{A}(p(\cdot), \omega(\cdot))$ contains both sides of the wall, and is therefore equal to (or nearly equal to) the set of maximally visible points $\mathcal{V}(p_0, p_1)$.  Therefore, the coverage $\mathcal{C}(p(\cdot), \omega(\cdot))$ will be equal to (or nearly equal to) $1$, as desired.
}

\subsection{Algorithmic Approach}
\label{sec:coverage_algorithm}

Given the above definition of coverage $\mathcal{C}(p(\cdot), \omega(\cdot))$, our goal is an algorithm which will compute the coverage given the video stream produced by the camera trajectory $(p(\cdot), \omega(\cdot))$.  We will use a deep learning pipeline for computing the coverage.  Unlike the case of depth estimation, our pipeline will need to be supervised, as there is no straightforward unsupervised loss that can be used.  Therefore, we need labelled training data; we now describe the data, following which we describe the general learning approach.

\revisedtag{13}{
\parbasic{Training Data} To gather training data, the most natural way to proceed would be to have physicians label video segments according to their coverage scores, and to use these labels as ground truth.  This is a standard approach to learning classification, detection, and segmentation models.  However, there is a problem with using this approach in the case of coverage: it turns out that physicians have considerable difficulty in accurately estimating coverage scores.  To illustrate these issues, we asked physicians to label synthetic video clips, and then we compared the physicians' labels with the ground truth.
}

\begin{figure}[t]
  \centering
  \begin{tabular}{cc}
    \fig{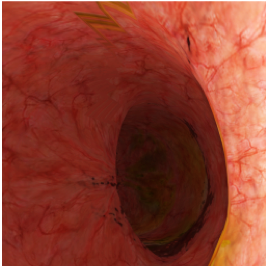}{3.0} & \fig{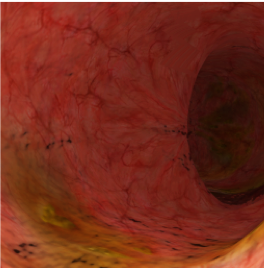}{3.0}
  \end{tabular}
  \caption{Example simulated images.}
  \label{fig:simulated_images}
\end{figure}

\revisedremove{As it is impossible to label real videos with coverage scores, we use synthetic videos.  Our}
More specifically, our videos are synthesized based on a colon simulator developed by 3D Systems \cite{3dSystems2019GIMentor}.  The colon is represented by a fully texture-mapped mesh, which can then be rendered using standard rendering engines; we have chosen to use Blender \cite{blender2019blender}.  Many different trajectories can be generated by taking a base trajectory and adding randomly chosen smooth curves in both the position and orientation of the camera; by doing so, we can generate many different full simulated colonoscopy procedures.  Each full procedure is then cut into short segments of 10 seconds, or 300 frames.  \revisedremove{Our ultimate goal is then to learn to compute coverage on a per-segment basis.}  Example simulated images are shown in Figure \ref{fig:simulated_images}.

\revisedtag{14}{
The advantage of using such synthetic video clips is that we have the actual ground truth for such clips: given the geometric model of the colon (from which we render the clips), we can compute the coverage.  Thus, the physicians' estimated coverage can easily be compared the ground truth.  We generated 561 such videos, and asked six physicians to label them; these physicians were gastroenterologists, with experience levels between four and ten years in gastroenterology, with a median of seven years.  We began by asking the physicians to label the actual coverage score, expressed
}
\revisedcont{
as a percentage.  Specifically, the task was explained to the physicians as ``express the percentage of actual surface viewed out of the total possible surface that could be viewed''; the task was carefully explained to each physician, along with some training examples.  The labels of the physicians, expressed as percentages, were then converted to fractions lying in $[0, 1]$.  When compared with the ground truth labels, physicians had a mean absolute error (MAE) of 0.177, which is very large.  The magnitude of this error is best viewed by examining the scatter plot of physicians' scores vs. ground truth labels, see the right subfigure of Figure \ref{fig:per_segment_scatter}: ideal performance would lie along the diagonal, but in practice the points are very far away from the diagonal.

We were interested to see whether physicians' performance on the labelling task was due to the fact that the label was expected to be a continuous variable (a percentage / fraction), and that such a task might be difficult or unnatural for many physicians.  We therefore gave the physicians a much simpler task, namely to decide whether in a given segment the colon was (1) ``mostly covered'', (2) ``partially covered'', or (3) ``mostly not covered''.  This 3-way classification task should be relatively straightforward.  There remained the issue of how to map ground truth coverage scores, which lie in $[0, 1]$ to these categories.  We therefore computed the mapping that maximized the physicians' accuracy on the task, i.e. that correlated best with the physicians' labels.  The result was equally convincing: on this 3-way classification task, physicians achieved an accuracy of $64.5\%$.  In fact, even when the ``partially covered'' and ``mostly not covered'' classes were combined, so that the classification task was now a binary task, the accuracy only increased slightly, to $67.6\%$.  These results show quite definitively that physicians have quite a difficult time estimating coverage.

We note that in general, the synthetic clips tend to be easier to label than real colonoscopy clips: the motion is smoother and slower, and there are fewer distracting artifacts (e.g. spraying of fluids).  Thus, we would assume that the conclusions drawn based on the above statistics would apply at least as much to real colonoscopy clips, and perhaps to an even greater degree.  The evidence provided above convinced us that we would need to train on synthetic videos, but to do so in such a way that we could then generalize to real videos.  We describe the manner in which we did this next.
}

\parbasic{The Algorithm} Our approach to learning coverage is to break the training regime into two separate stages.  \revisedtag{15}{
We begin by noting that a special case of coverage can be computed when only a single frame is considered.  In this case, the trajectory is just a single pose $p, \omega$, so that coverage as defined in Equation (\ref{eq:coverage}) becomes
\[
\mathcal{C}_{single}(p, \omega) = \frac{\mu[\mathcal{A}(p, \omega)]}{\mu[\mathcal{V}(p, p)]}
\]
Note from Equations (\ref{eq:maximally_visible}) and (\ref{eq:actually_visible}) that both $\mathcal{V}$ and $\mathcal{A}$ are well-defined for a single frame.  In the first stage, then, we train a per-frame network, whose input is a single depth image, and whose output is the coverage for that frame $\mathcal{C}_{single}(p, \omega)$.  In practice, we use a vector of outputs of several coverages, each computed with different viewing angle and look-ahead parameters, corresponding to $\Delta_0$ and $\Delta_1$ in Equation (\ref{eq:maximally_visible}).}   In the second stage, we strip the final layer off of the per-frame network, exposing the penultimate layer which is then taken as a feature vector.  We then train a per-segment coverage network by taking as input the collection of feature vectors, one for each frame in the segment; and the output is the segment coverage $\mathcal{C}(p(\cdot), \omega(\cdot))$.  The structure of the two stage procedure is shown in Figure \ref{fig:network_structure}.

Why proceed with a two stage procedure, rather than a single stage?  There are three primary reasons:
\begin{itemize}
    \item \textbf{Allows for Simple Domain Adaptation:} We are training on synthetic videos, but the ultimate goal is for the networks to predict coverage on real videos.  The concern is that the networks -- given their large capacity -- may learn to predict coverage based on some minor artifacts of the synthetic videos which do not then generalize to real videos.
    \revisedremove{The two stage approach enables domain adaptation, as }
    To deal with this, the initial per-frame network learns a feature representation based on a rather coarse representation of the 3D geometry, namely the visibility.  Only this feature vector is then used in the final per-segment network.
    \revisedtag{16}{
    Due to the geometric coarseness of these features, a very simple domain adaptation scheme may be employed.  In particular, we use the ``frustratingly easy'' technique of Sun \etal \cite{sun2016return}, which applies an affine transformation to the second last layer of the per-segment network.  This affine transformation causes the mean and covariance of this layer's output, computed over the real segments, to be transformed to match the corresponding statistics for the synthetic segments.  This very simple domain adaptation technique is all that is required to achieve high performance on real segments, as we show in Section \ref{sec:results_coverage}.
    }
    \item \textbf{Less Training Data:} Synthesizing full videos is a rather heavy operation, as each frame must be rendered, and a video of 5 minutes consists of 9,000 frames.  The natural approach, which learns coverage directly from video segments, would require many such segments to converge; and this would necessitate the rendering of a very large number of frames.  Using the two stage approach mitigates this problem: a modest number of video segments, on the order of hundreds, will  still consist of many frames.  The per-frame network will therefore have a lot of data to learn from (hundreds of thousands of images); whereas the per-segment network will be learning a much easier task, and can therefore learn on a considerably smaller amount of data (hundreds of segments) using a network with much lower capacity.
    \item \textbf{Inference Speed:} A natural candidate for the architecture of a direct approach is a 3D CNN; this is the standard architecture that is used in action recognition, for example \cite{carreira2017quo}.  Unfortunately, such networks are quite heavy, and cannot generally run in real-time.  Other approaches for spatio-temporal data include combined recurrent-convolutional architecture \cite{carreira2017quo,liu2018mobile}.  Our proposal, by contrast, is a straightforward convolutional architecture, which is very clean and easy to train.
\end{itemize}

\revisedtag{17}{
It is natural to wonder whether one can combine the two stages into a single, via a unified loss function.  We note that although this might be possible, there is a distinct advantage to keeping the training of the two stages separate, due to the fact that we have many more frames than we have segments.  The per-frame model that is learned is therefore quite accurate, as it is trained on a very large number of examples.  Once the per-frame model has been learned, the per-segment model can benefit from the per-frame model via the use of the representation that has been learned in the per-frame model.  This enables a kind of transfer learning, which is very useful since there are far fewer segments than there are frames.  If one were to learn on both frames and segments simultaneously, it is not clear if this transfer learning would work as well.  For example, the part of the loss related to the segments might ``drown out'' the part of the loss related to the frames, which would be problematic, given the relatively small number of segments.
}

\subsection{Network Architecture}

\begin{figure}[t]
  \centering
  \fbox{\fig{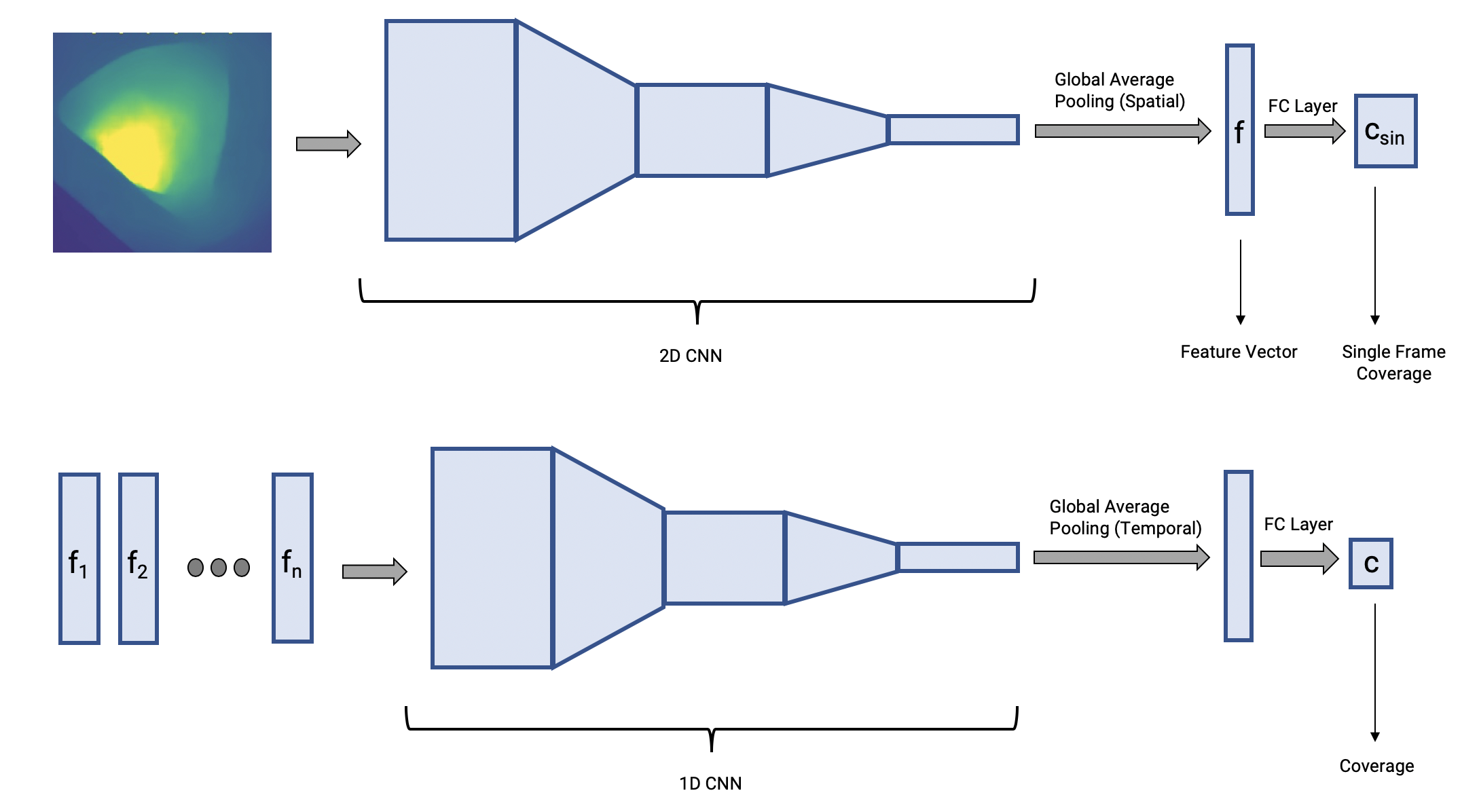}{8.5}}
  \caption{Network structure.  See accompanying description in the text.}
  \label{fig:network_structure}
\end{figure}

We begin with the per-frame network.  The input to the per-frame network is a depth image.  We use a ResNet-50 architecture, strip off the final layer, and replace it with a fully connected layer which reduces to a vector of size three.  Each entry of this vector is the visibility computed for different parameters $(\Delta_0, \Delta_1)$.  This is trained with an $L_2$ loss.  See Figure \ref{fig:network_structure} for an illustration of the per-frame network structure.  After training is complete, we strip off the last layer of the per-frame network, so that the new output (previously, the penultimate layer) is a feature vector of length 2,048.

We now turn to the per-segment network.  A segment is taken to be 10 seconds worth of video, which at 30 fps translates to 300 frames.  Each frame is passed through the per-frame network, yielding a collection of 300 vectors, each of length 2,048.  We consider this to be a 2-tensor, of length 300, and with 2,048 channels.  This 2-tensor is then the input to the network, which is a 1D CNN.  This network is relatively small, as we have a small number of training samples; there are six 1D convolutional layers, followed by average pooling over the temporal dimension and a final fully-connected layer (see Figure \ref{fig:network_structure}). The total number of parameters of the network is 20K, which is quite small.  

In practice, run-time inference proceeds as follows.  The current RGB frame is passed through the depth estimation network described in Section \ref{sec:depth}.  This depth image is then passed through the per-frame network, producing a vector of length 2,048.  This vector is then appended to the vectors from the previous 299 frames (which were computed at previous time steps, and stored in memory) to create the 2D tensor which is then passed into the per-segment network.  \revisedremove{The coverage operations are both light, and can be achieved in real-time.}
\revisedtag{18}{
Regarding timing, we can compute the overall time required on a Titan Xp GPU using the numbers measured in \cite{bianco2018benchmark}, and scaled by the change in resolution (from $224 \times 224$ as used in \cite{bianco2018benchmark} to $384 \times 320$).  The depth network uses a ResNet-18 architecture, which requires 4.38 ms, while the per-frame network uses a ResNet-50 architecture which requires 12.49 ms.  The per-segment network uses a non-standard architecture, which is nevertheless quite light; our own timing experiments indicate that it runs in 0.20 ms on a CPU, and therefore would require less on a GPU.  The total time for all stages is therefore less than 17.07 ms, which would allow for frame-rates of up to 58 frames per second.
}

\section{Results}
\label{sec:results}
\subsection{Depth Estimation}

\parbasic{Description of the Dataset}  We have three different sources of data for evaluating our depth estimation algorithm.  The first source is the dataset introduced in \cite{rau2019implicit}, which we refer to as the UCL dataset.  This is a synthetic dataset, for which ground truth depth is available, consisting of 16,016 (RGB, depth) image pairs, with a train-test split of 10,556 vs. 5,460.  The second source is based on synthetic videos we have generated, using the colon simulator developed by 3D Systems \cite{3dSystems2019GIMentor}, which were then rendered using Blender \cite{blender2019blender}.  Again, ground truth is available for this set, which consists of 187,369 (RGB, depth) image pairs, with a train-test split of 134,025 vs. 53,344.  We refer to this dataset as the Google-Synthetic dataset.  The final source is real de-identified colonoscopy videos from Orpheus Medical, which have been recorded at 16 mbps; we have acquired 3,049 such videos.  These are very useful for training, given that our algorithm is unsupervised, but cannot be used for quantitative evaluation as no ground truth is available.  We refer to this dataset as the Google-Real dataset.

\parbasic{Metrics} We report a few metrics for the quality of the depth estimation.  The first metric is Mean Relative Error (MRE).  In the ordinary way, MRE would be defined as
\begin{equation}
    \text{MRE} = \frac{1}{n} \sum_{i=1}^n \frac{|\hat{d}_i - d_i|}{d_i}
    \label{eq:mre_standard}
\end{equation}
where $d_i$ is the ground truth depth of pixel $i$, $\hat{d}_i$ is the estimated depth, and the sum is over all the pixels in an image (or in multiple images).  However, note the fact that the depth estimation algorithm is only correct up to scaling, so we must take this into account; furthermore, the above formula does not account for the case where the actual ground truth depth is $0$.  Thus, we emend the formula in (\ref{eq:mre_standard}) to read
\begin{equation}
    \text{MRE} = \min_\sigma \frac{1}{n} \sum_{i=1}^n \frac{|\sigma \hat{d}_i - d_i|}{\max(d_i, \epsilon)}
    \label{eq:mre_scale_insensitive}
\end{equation}
The minimization allows us to choose the best scaling parameter for the test set, and the term in the denominator accounts for when the ground truth depth is $0$.

Our hypothesis is that much of the error in the depth reconstruction comes near discontinuities.  Specifically, if a discontinuity is correctly reconstructed, but its position is off by one pixel, then the MRE incurs a large loss.  A measure closely related to the MRE is the Discontinuity Robust MRE, or DR-MRE:
\[
\text{DR-MRE} = \min_\sigma \frac{1}{n} \sum_{i=1}^n \min_{j \in \mathcal{N}_i} \frac{|\sigma \hat{d}_i - d_j|}{\max(d_j, \epsilon)}
\]
where $\mathcal{N}_i$ is the $3 \times 3$ neighbourhood around pixel $i$.  If the position of the discontinuity is off by a single pixel, then the DR-MRE will be insensitive to this.

Finally, another way of dealing with the scaling issue is to note that order is preserved under scaling.  Therefore, we may also verify whether the order of pairs of pixels is the same for both the ground truth and estimated depths.  Given a pair of pixels, we define $r_{ij} = \mathbb{I}[d_i > d_j]$, where $\mathbb{I}[\cdot]$ is the indicator function; and similarly, $\hat{r}_{ij} = \mathbb{I}[\hat{d}_i > \hat{d}_j]$.  Then we define the Depth Order Measure (DOM) to be
\[
\text{DOM} = \frac{1}{|\mathcal{P}|} \sum_{(i, j) \in \mathcal{P}} \bigg[ r_{ij} \hat{r}_{ij} + (1-r_{ij})(1-\hat{r}_{ij}) \bigg]
\]
where the sum is taken over all pairs of pixels.  In practice, the number of pairs is enormous, so a sampling strategy is employed.  Note that the above measure is similar to the Rand Index.

\parbasic{Results} The performance of the depth estimation algorithm is reported in Table \ref{tab:depth}.  The numbers for the MRE indicate that on average, the estimated depth is within 5.2\% or 16.8\% of the true value for the Google-Synthetic and UCL datasets, respectively.  However, examining the DR-MRE, we see that in the case of the UCL dataset, more than half of the error is due to small errors in discontinuity placement: the DR-MRE is only 7.9\%.  The gains for the Google-Synthetic dataset are considerably more modest in going from MRE to DR-MRE, indicating that the algorithm was better at placing depth discontinuities for this dataset.
\revisedtag{19}{
There are two potential reasons for this.  First, Google-Synthetic is more than 10 times larger than UCL: the train set sizes are 134,025 vs. 10,556.  Thus, there may have simply been enough data to learn the discontinuities better.
Second, Google-Synthetic appears to be somewhat smoother than UCL.
}

\revisedtag{20}{
We also note that Rau \etal \cite{rau2019implicit} achieve MRE = 6.4\% on the UCL dataset; this value is computed using
}
\revisedcont{
the standard definition of MRE in Equation (\ref{eq:mre_standard}), rather than the scale insensitive version in (\ref{eq:mre_scale_insensitive}).  However, note that the algorithm of Rau \etal is fully supervised, whereas ours is completely unsupervised.
}

We now examine the DOM values.  The DOM lies in $[0,1]$, and is quite high at $0.978$ and $0.933$ for the Google-Synthetic and UCL datasets, respectively.   This indicates that order is preserved nearly all of the time.  Our hypothesis is that such numbers would be sufficient to enable the coverage algorithm; this will be borne out in Section \ref{sec:results_coverage}.  
\begin{table}[h]
\centering
\begin{tabular}{|c||c|c|c|}
  \hline
    & MRE & DR-MRE & DOM  \\
  \hline \hline
  Google-Synthetic & 0.052 & 0.046 & 0.978 \\
  \hline
  UCL \cite{rau2019implicit} & 0.168 & 0.079 & 0.933 \\
  \hline
\end{tabular}
\caption{Performance of the depth estimation algorithm.}
\label{tab:depth}
\end{table}

\begin{figure}[t]
  \centering
  \begin{tabular}{cccc}
    \hnb \fig{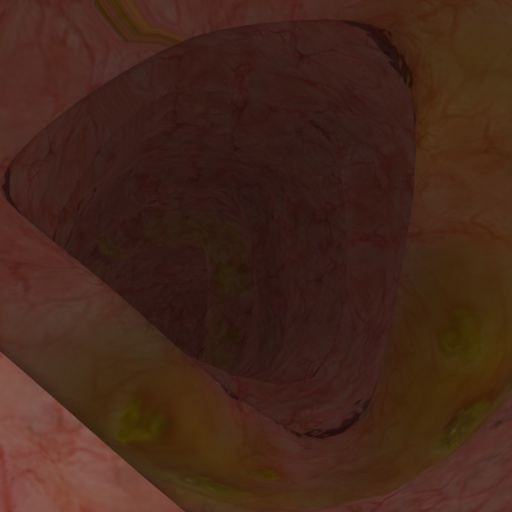}{2.1} & \hn \fig{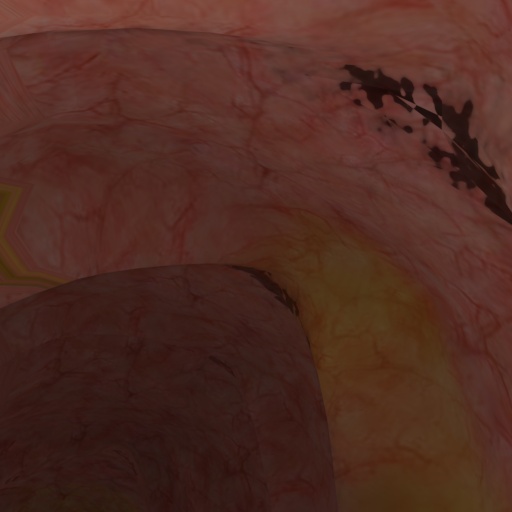}{2.1} & \hn \fig{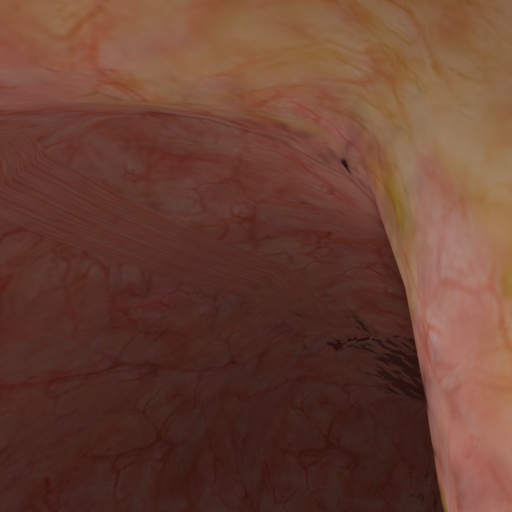}{2.1} & \hn \fig{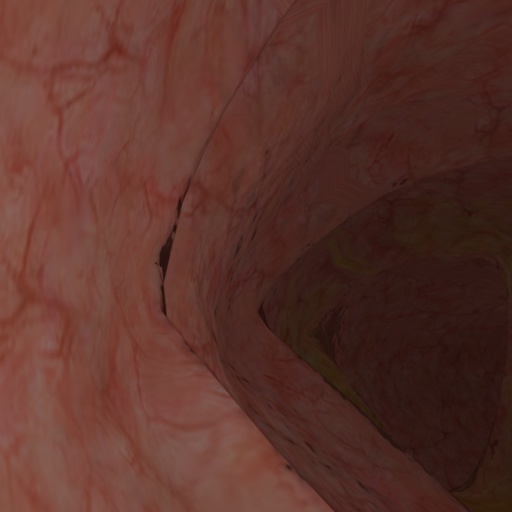}{2.1}  \\
    \hnb \fig{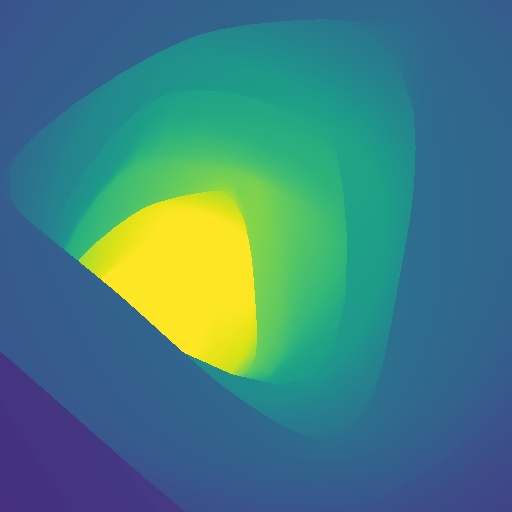}{2.1} & \hn \fig{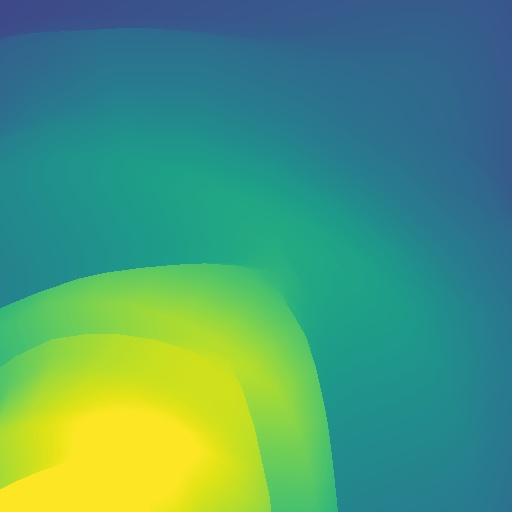}{2.1} & \hn \fig{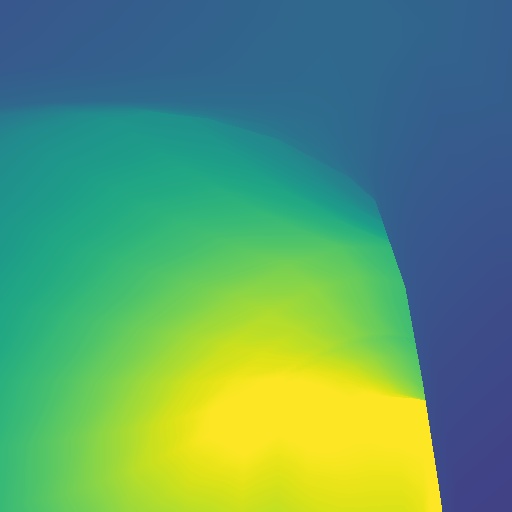}{2.1} & \hn \fig{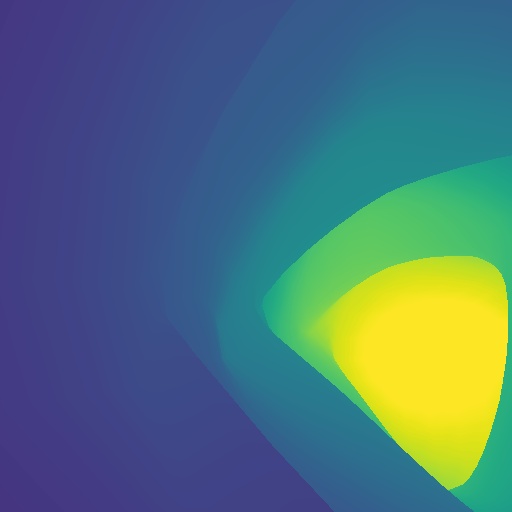}{2.1}  \\
    \hnb \fig{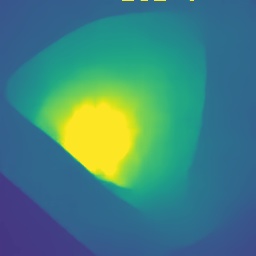}{2.1} & \hn \fig{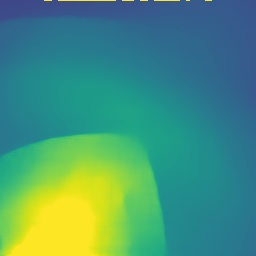}{2.1} & \hn \fig{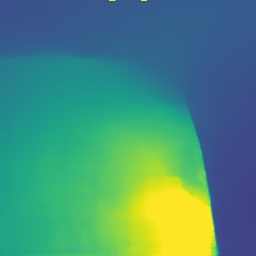}{2.1} & \hn \fig{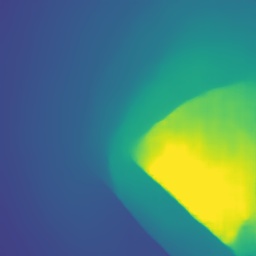}{2.1}
  \end{tabular}
  \caption{Depth estimation on the Google-Synthetic dataset.  Top: RGB image.  Middle: ground truth depth map.  Bottom: estimated depth map.  Yellow is deeper, blue is more shallow.}
  \label{fig:depth_images_synthetic}
\end{figure}

\begin{figure}[t]
  \centering
  \begin{tabular}{cccc}
    \hnb \fig{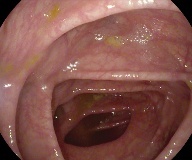}{2.1} & \hn \fig{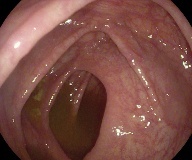}{2.1} & \hn \fig{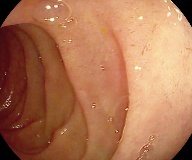}{2.1} & \hn \fig{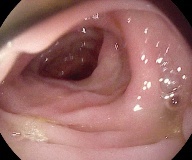}{2.1}  \\
    \hnb \fig{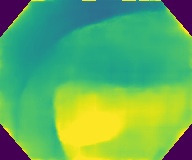}{2.1} & \hn \fig{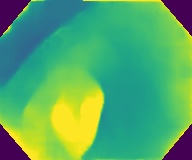}{2.1} & \hn \fig{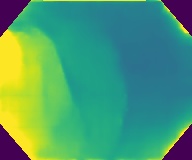}{2.1} & \hn \fig{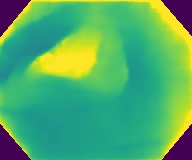}{2.1}
  \end{tabular}
  \caption{Depth estimation on the Google-Real dataset.  Top: RGB image.  Bottom: estimated depth map.  Yellow is deeper, blue is more shallow.}
  \label{fig:depth_images_real}
\end{figure}

We now turn to more qualitative results, by showing example depth maps from the various datasets; this is the natural way of judging the quality of the algorithm on the real dataset (given the lack of ground truth), but it also gives a better flavor of the performance on synthetic datasets.  Synthetic results on the Google-Synthetic dataset are shown in Figure \ref{fig:depth_images_synthetic}; note the striking similarities between the ground truth depth maps and their estimated counterparts.  Real results on the Google-Real dataset are shown in Figure \ref{fig:depth_images_real}.  One can see the overall features of the colon are captured nicely, including the depth of the ``tunnel'' down the lumen, as well as the folds.

\subsection{Coverage}
\label{sec:results_coverage}

\parbasic{Description of the Dataset}  The dataset comes from two sources.  The first source consists of synthetic videos which were generated using the colon simulator developed by 3D Systems \cite{3dSystems2019GIMentor}, and then rendered using Blender \cite{blender2019blender}.  These videos are then divided into segments of 10 seconds duration, i.e. 300 frames; in total, 561 such video segments were generated.  Each of these segments possesses a ground truth coverage label in $[0,1]$.  Note that in experiments, we use 5-fold cross-validation, allowing us to test on all 561 sequences.  In addition, each such video was given a coverage value by a gastroenterologist, which allows us to compare C2D2's performance to that of human experts.  \revisedtag{21}{This setup has already been discussed at length in Section \ref{sec:coverage_algorithm}.  We are releasing this dataset, which is located at [TBD REPOSITORY].}

The second source consists of real videos, which are full colonoscopy procedures which have been recorded at 16 mbps; this is the Google-Real dataset, which has already been described in the context of depth estimation.  As in the case of the synthetic videos, the real videos are divided into segments, which are randomly chosen 10 seconds subsequences.  These segments do not possess a ground truth label.
\revisedremove{
As we shall see shortly, human experts are not particularly accurate in estimating coverage (they have a high Mean Average Error); thus, one cannot use physicians to provide ground truth labels for real sequences.  As a result, it is not possible to quantitatively assess performance on real sequences.  Nevertheless, we can provide \emph{qualitative} performance results on real sequences, verifying the algorithm's output with the eyeball test.
}
\revisedtag{22}{
As discussed at length in Section \ref{sec:coverage_algorithm}, human experts are not particularly accurate in estimating coverage -- either in terms of the MAE of coverage, or accuracy on simpler classification tasks.  Thus, one cannot use physicians to provide ground truth labels for real sequences.  Instead, for quantitative validation on real segments we use an ``expert verification'' scheme, which we detail shortly, in the discussion concerning per-segment results.
}

\begin{figure}[ht]
  \centering
  \begin{tabular}{cc}
    \fig{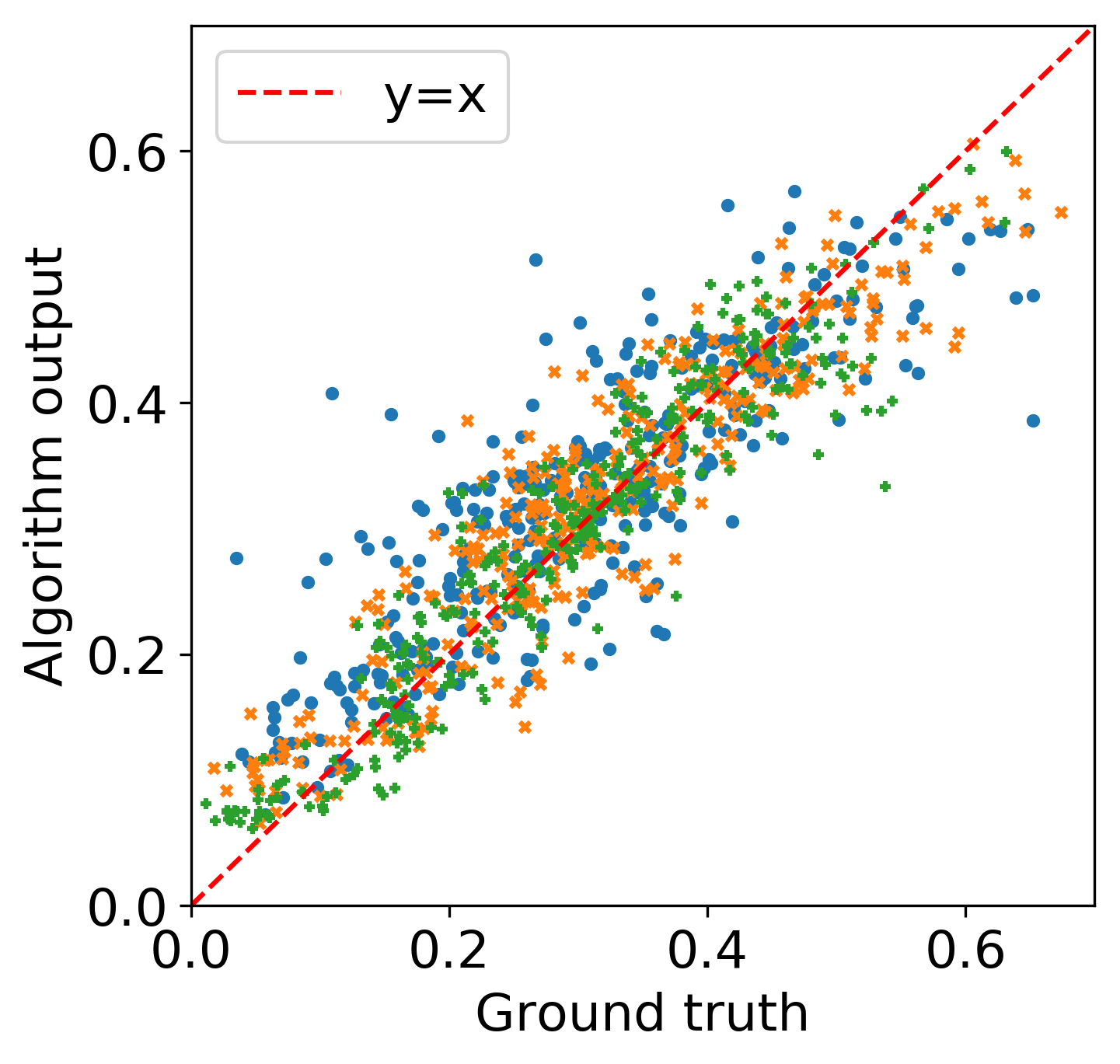}{5.0} & \raisebox{1.3cm}{\fig{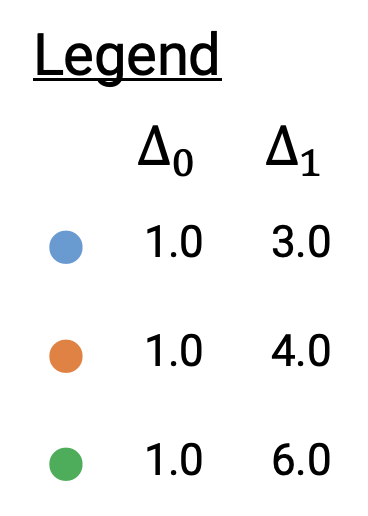}{2.0}}
  \end{tabular}
  \caption{\captiontag{24} \revised{Scatter plot of the predicted single frame coverage vs. true single frame coverage.  The color denotes different parameter values for the visibility computation, see Section \ref{sec:coverage}.}}
  \label{fig:per_frame_scatter}
\end{figure}

\parbasic{Results: Per-Frame} We begin by describing results for the per-frame network.  Although the per-frame task does not represent our final goal, it is nevertheless interesting to report results on this first stage of our two stage mechanism.  
\revisedtag{23}{
We use three different values for the pair $(\Delta_0, \Delta_1)$, namely $(1.0, 3.0)$, $(1.0, 4.0)$, and $(1.0, 6.0)$.
We first show a scatter plot of the predicted single frame coverage vs. true single frame coverage, for the test set in Figure \ref{fig:per_frame_scatter}; each color corresponds to a different value of $(\Delta_0, \Delta_1)$ as shown in the legend.  As can be seen, the network succeeds in learning, fairly well, how to predict single frame coverage.
}

\begin{figure}[ht]
  \centering
  \fig{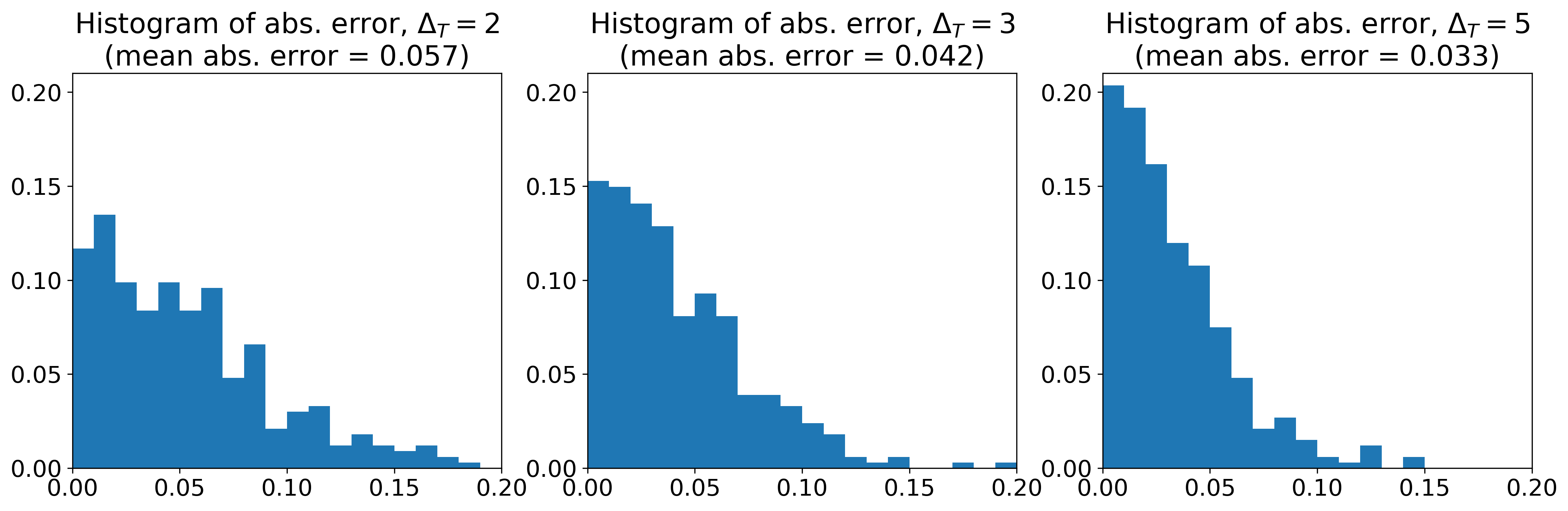}{9.0}
  \caption{\captiontag{25} \revised{Histograms of Mean Absolute Error (MAE) of the per-frame visibility network, for three separate parameter pairs $(\Delta_0, \Delta_1)$. Left: $(1.0, 3.0)$; middle: $(1.0, 4.0)$; right: $(1.0, 6.0)$.}}
  \label{fig:per_frame_mae}
\end{figure}

A more quantitative measure of the performance is the Mean Absolute Error (MAE), which we report separately for each of the parameter values, shown in Figure \ref{fig:per_frame_mae}.  The MAEs range between $0.033$ and $0.057$, which is quite reasonable.  It is crucial to remember that the actual performance is immaterial in the end; rather, this network is used as a feature extractor for the ultimate goal, which is per-segment coverage.  Nonetheless, it is important that the feature extractor be informative, and achievement of the intermediate goal of single frame coverage prediction indicates that this is indeed the case.

\begin{figure}[t]
  \centering
  \begin{tabular}{cc}
    \fig{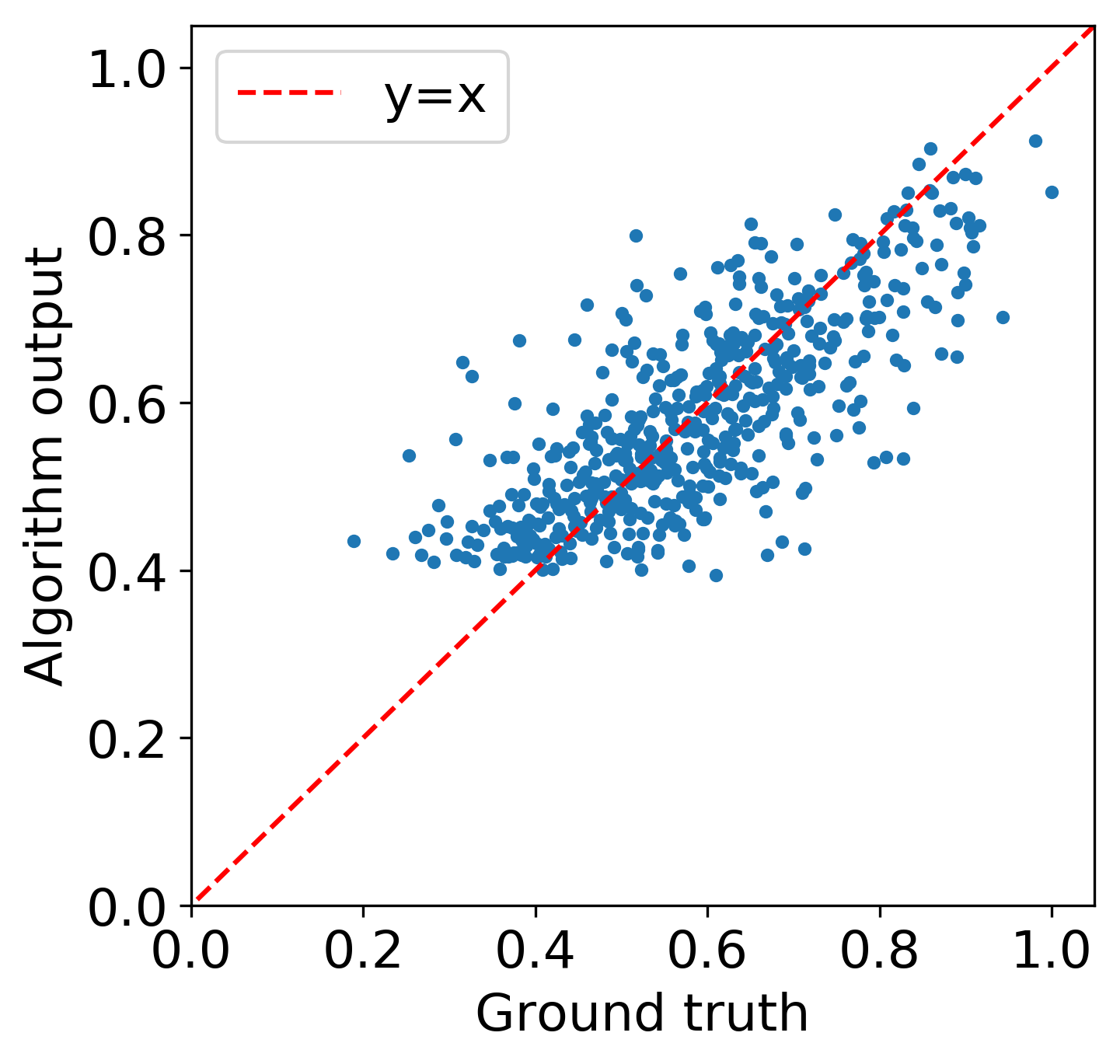}{4.0} & \fig{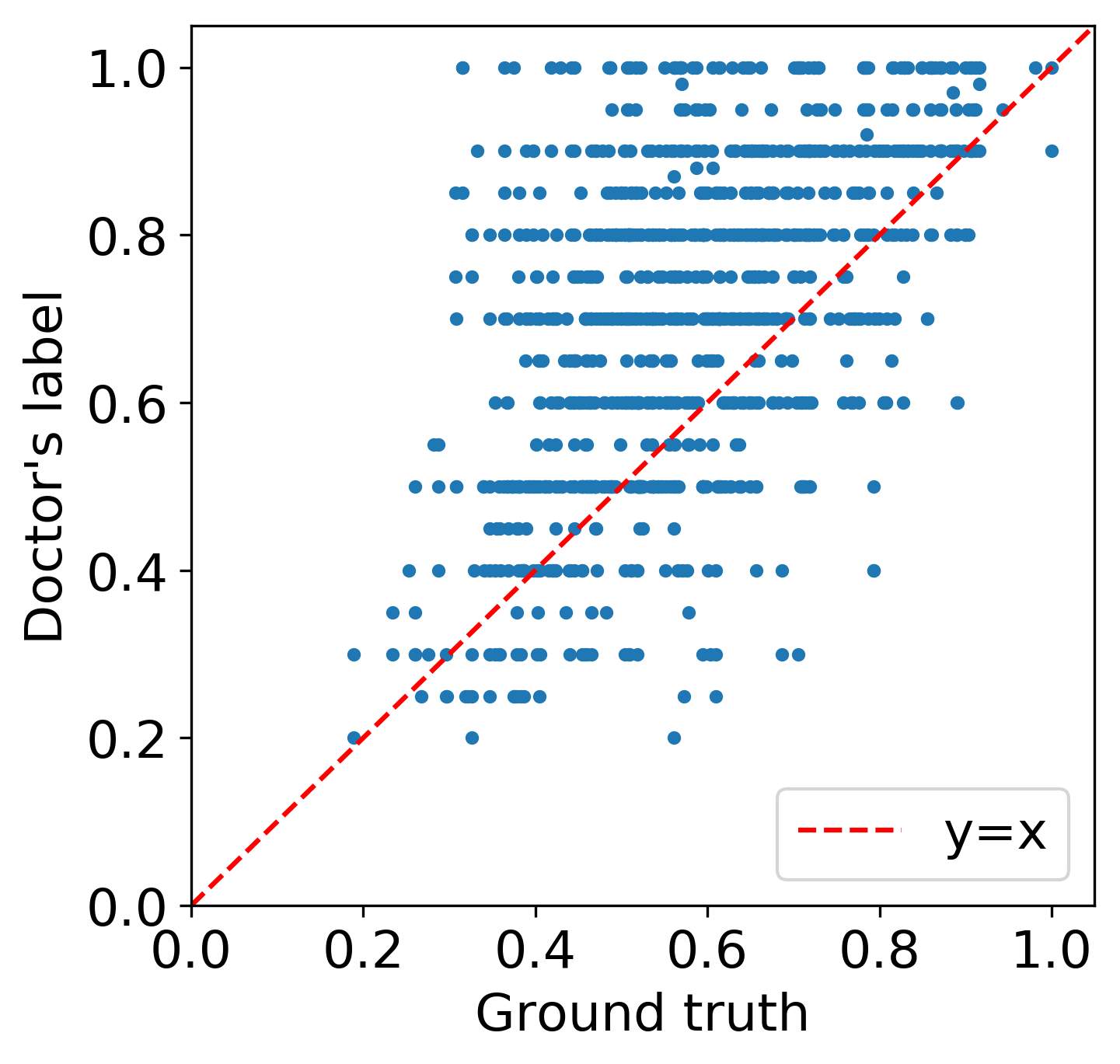}{4.0} 
  \end{tabular}
  \caption{Scatter plots of the predicted coverage vs. true coverage on synthetic sequences.  Left: C2D2's performance.  Right: physicians' performance.  C2D2's performance is considerably better.}
  \label{fig:per_segment_scatter}
\end{figure}

\parbasic{Results: Per-Segment}
We begin with a discussion of C2D2's performance on synthetic video segments, for which we have ground truth.  As in the previous section, we show the scatter plots of the predicted coverage vs. true coverage.  In Figure \ref{fig:per_segment_scatter}, the left plot shows C2D2's performance, while the right plot shows the physicians' performance.  Ideal performance entails clustering on the diagonal; as can be seen, C2D2's performance is considerably better than that of the physicians.

\begin{figure}[t]
  \centering
  \begin{tabular}{cc}
    \fig{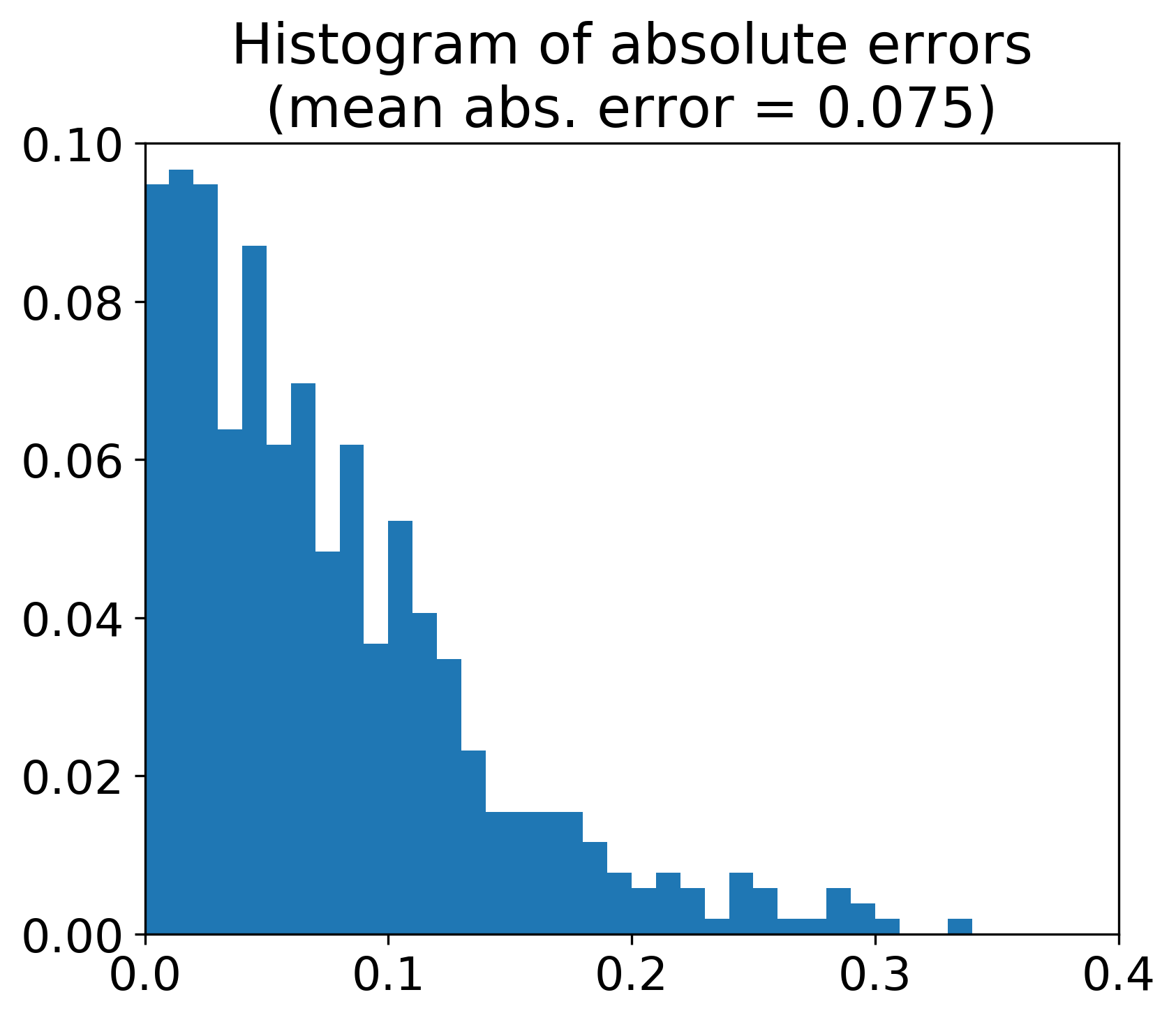}{4.2} & \fig{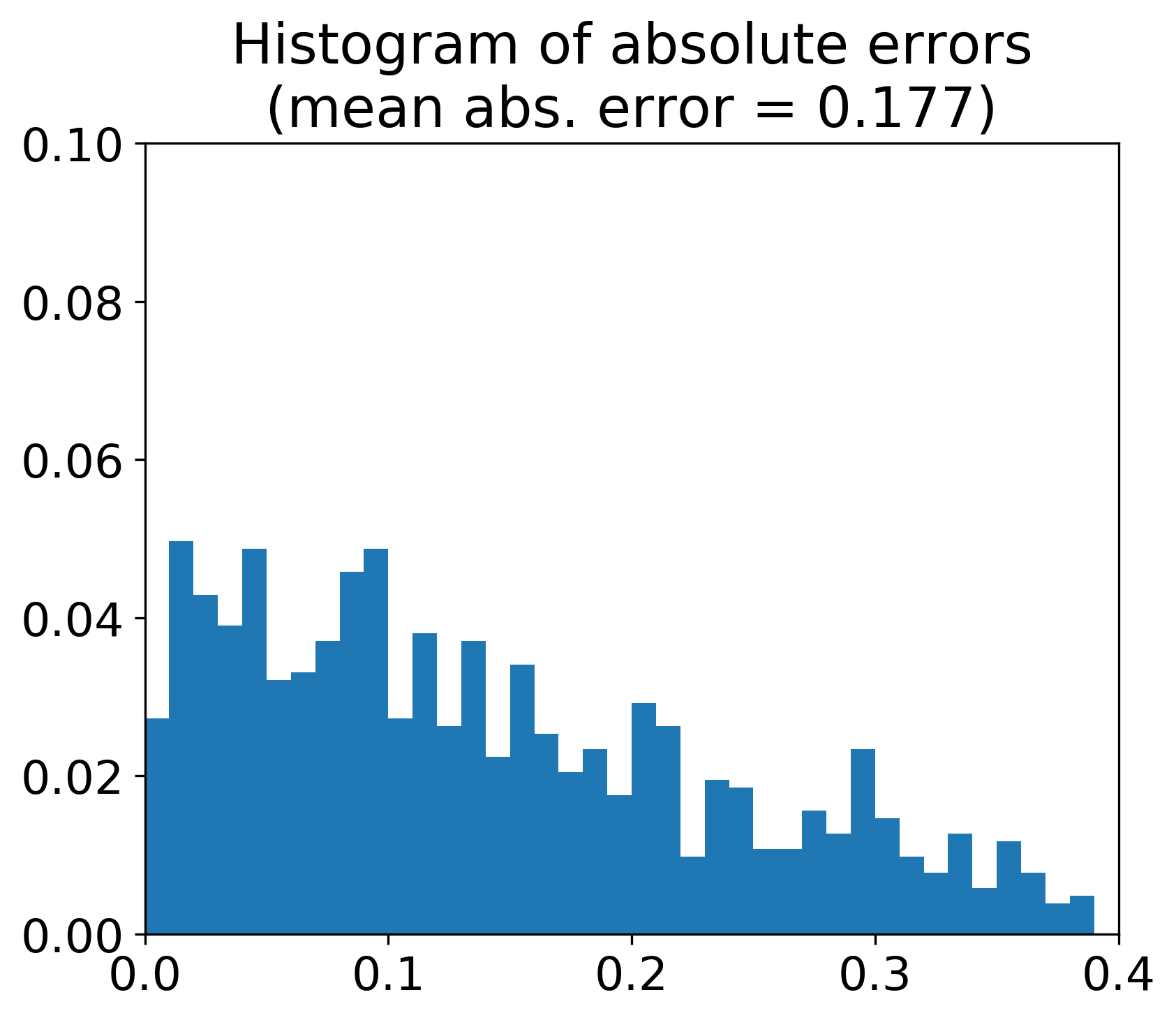}{4.0} 
  \end{tabular}
  \caption{Histograms of Mean Absolute Error (MAE) of coverage on synthetic sequences.  Left: C2D2's performance, MAE = 0.075.  Right: physicians' performance, MAE = 0.177.  C2D2's performance is 2.4 times better.}
  \label{fig:per_segment_mae}
\end{figure}

To further quantify the difference in performance, we examine the MAE for both C2D2 and the physicians, see Figure \ref{fig:per_segment_mae}.  C2D2 attains MAE = 0.075, while the physicians receive MAE = 0.177.  By this metric, C2D2's performance is 2.4 times better, clearly demonstrating the system's ability to outperform human experts.  

\revisedtag{26}{
We now turn to quantitative performance on real video segments.  As noted above, physicians have difficulty with labelling coverage on synthetic sequences (which tend to be easier than real sequences), incurring an MAE of 0.177.  Furthermore, as described in depth in Section \ref{sec:coverage_algorithm}, physicians also have difficulty with the simpler scenario of labelling sequences according to a 3-way classification task, in which the goal is to decide whether in a given segment the colon was (1) ``mostly covered'', (2) ``partially covered'', or (3) ``mostly not covered''.  In particular, the physicians' accuracy on this task was $64.5\%$.  Thus, we cannot simply have physicians label video clips according to these 3 classes and compare C2D2's predictions to these classes, as the physicians' labels are far too noisy.

Instead, we use a technique which is a variant of that used in the generative modelling literature \cite{salimans2016improved}: we ask the physicians to judge the algorithm's result.  More specifically, we present 
}
\revisedcont{
the physician with both the video segment, as well as C2D2's output, mapped to one of the three classes mentioned above; the physician is then asked whether they ``agree'' or ``disagree'' with C2D2's prediction.  In order to enable this task, we must have a way of mapping C2D2's coverage score in $[0, 1]$ to the three classes.  We did this by examining a small number of clips -- prior to the physicians' annotation -- and deciding on a sensible set of bins by eyeballing.  The bins were taken to be $[0, 0.4)$, $[0.4, 0.8)$, and $[0.8, 1]$.

The results of this performance evaluation are shown in Table \ref{tab:coverage}.  Two of the six physicians mentioned in Section \ref{sec:coverage_algorithm} (with 7 and 4 years experience as gastroenterologists) were given 385 clips; each clip was examined by a single physician.  The physicians rejected 84 of the clips as not relevant for the coverage task -- due to the presence of spraying of fluids, lack of motion due to a polyp being removed, etc.  This left 301 clips, in which there was agreement on 280.  Thus, in total there was physician agreement with C2D2's prediction on 93.0\% of the clips, which speaks to the accuracy of the algorithm.
\begin{table}[h]
\centering
\begin{tabular}{|p{1.5cm}|p{3.0cm}|p{3.0cm}|}
  \hline
    Total \# Clips & \# Clips with Physician Agreement with C2D2 & Percent Physician Agreement with C2D2  \\
  \hline \hline
  \centering 301 & \centering 280 & \hspace{1.0cm} 93.0\% \\
  \hline
\end{tabular}
\caption{Performance of C2D2 on real sequences.}
\label{tab:coverage}
\end{table}
}

\begin{figure}[t]
  \centering
  \begin{tabular}{ccccc}
  \hnb \rotatebox{90}{\hspace{0.35cm} \scriptsize{0.931}} & \hn \fig{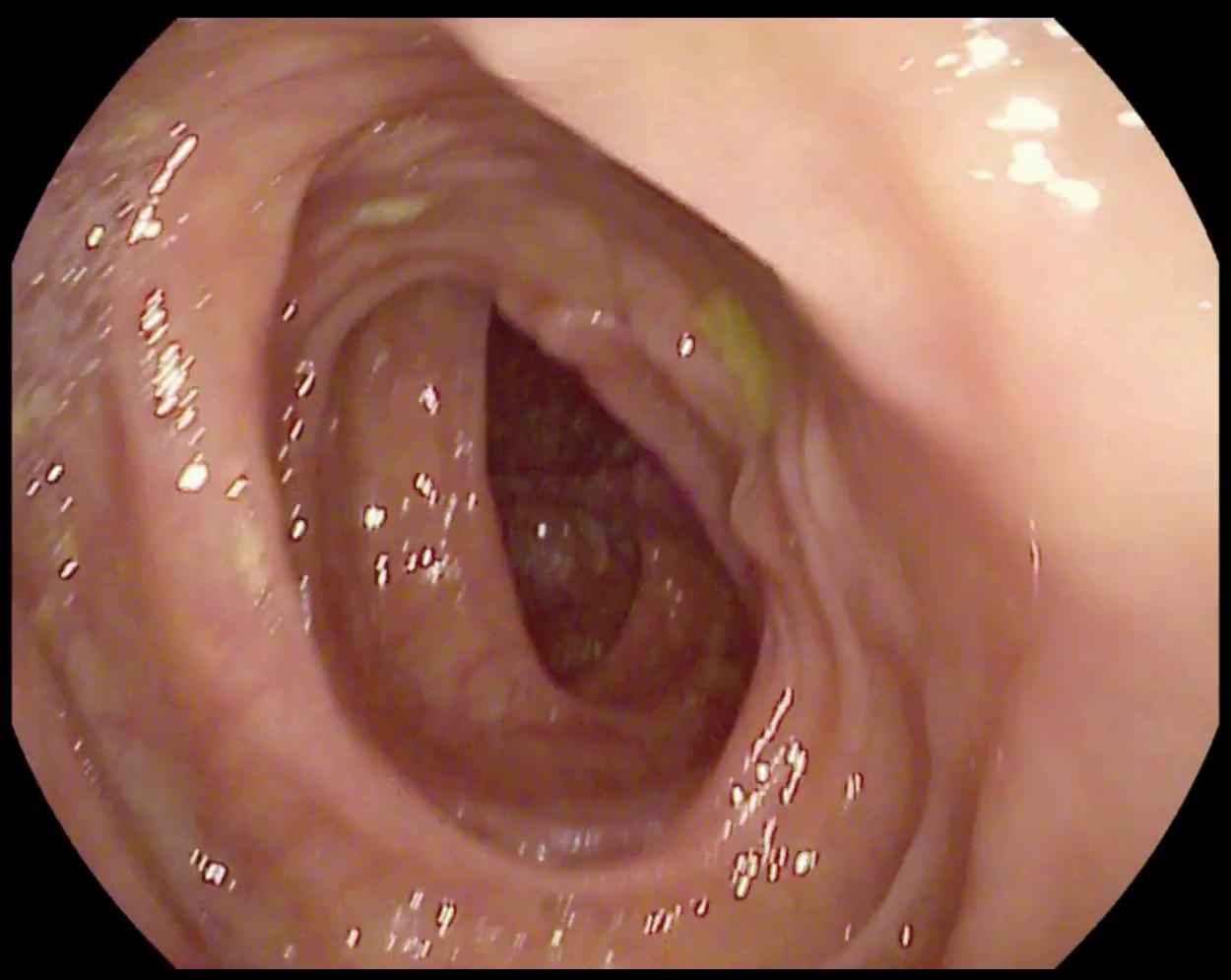}{2.0} & \hn \fig{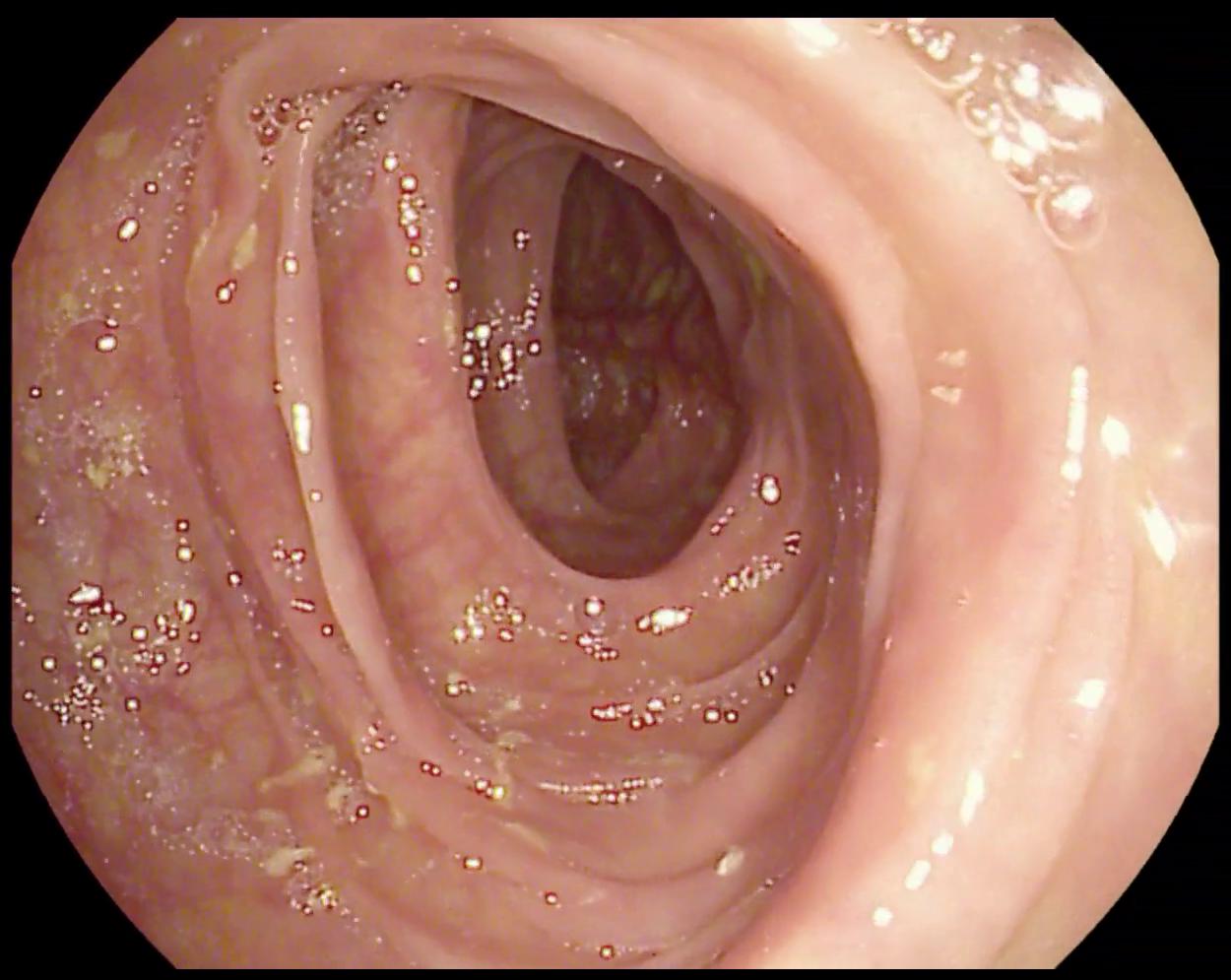}{2.0} & \hn \fig{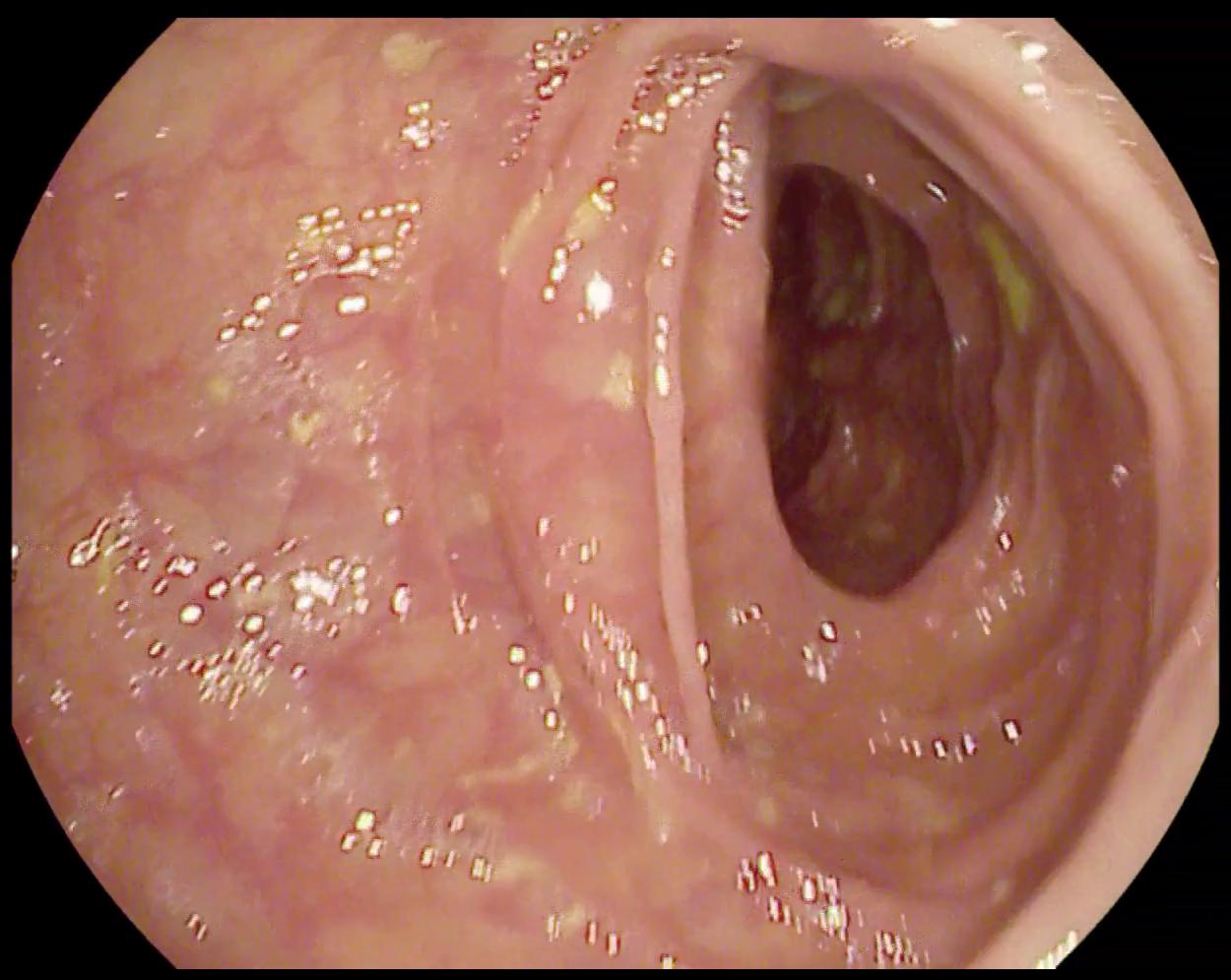}{2.0} & \hn \fig{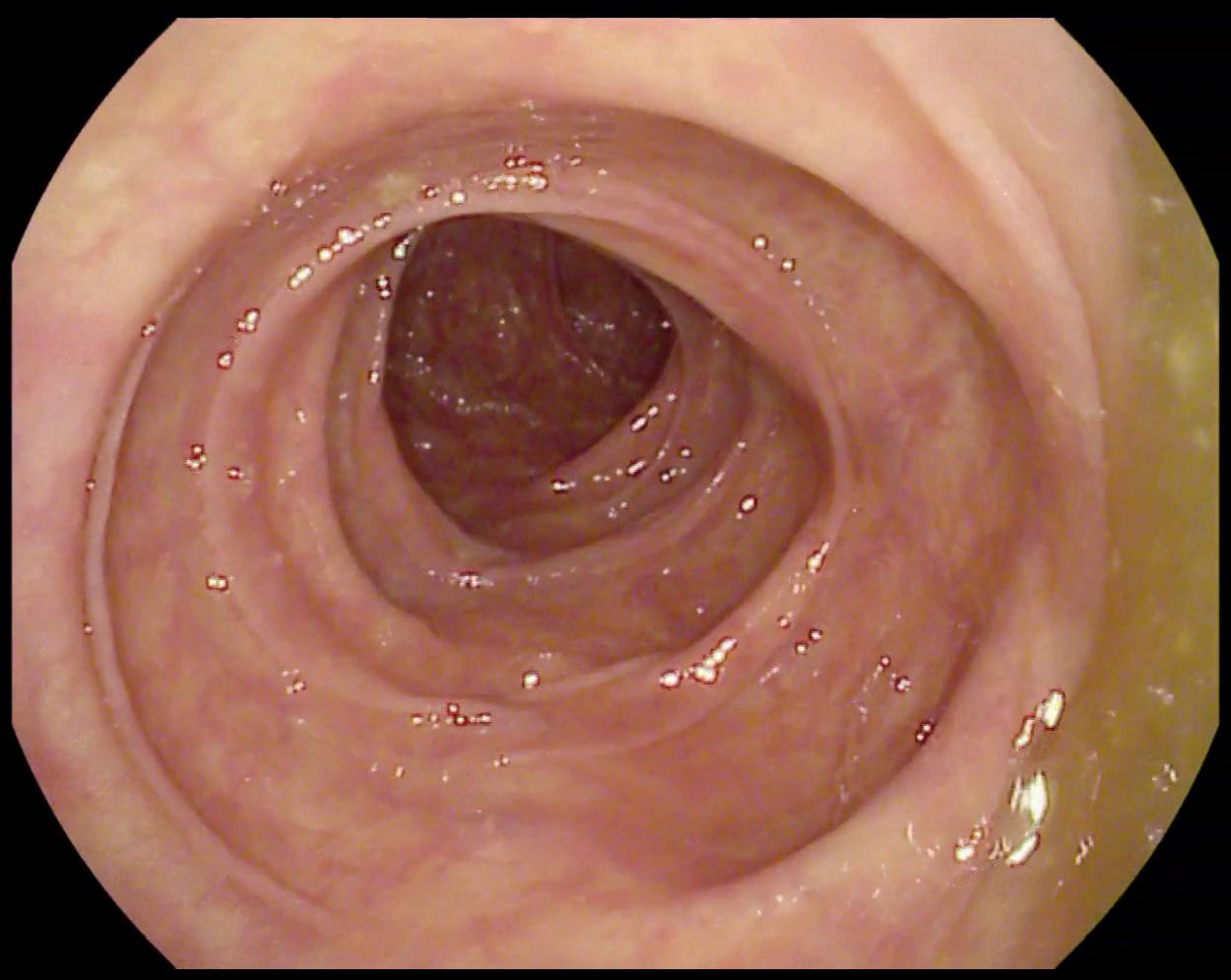}{2.0} \\
  \hnb \rotatebox{90}{\hspace{0.35cm} \scriptsize{0.920}} & \hn \fig{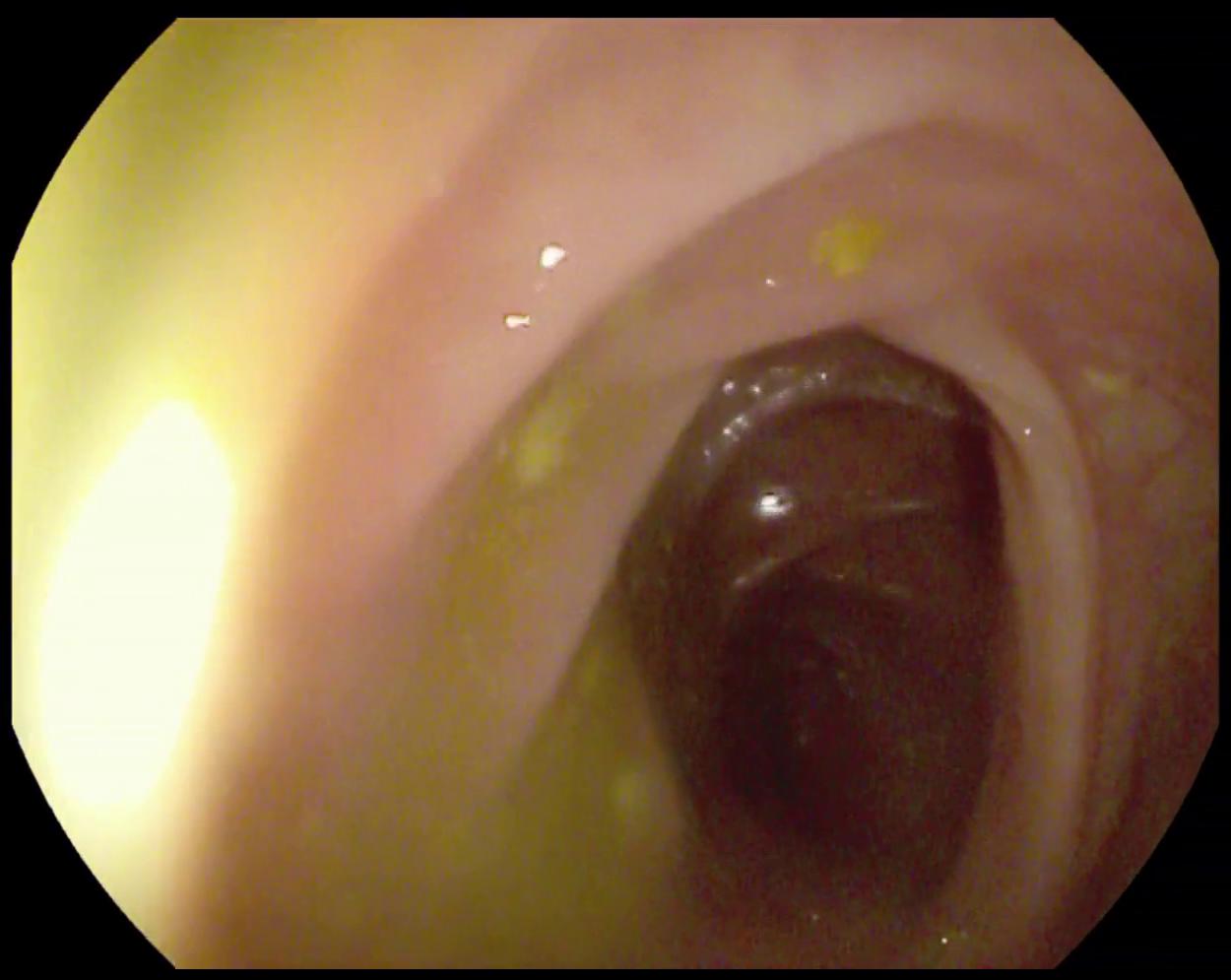}{2.0} & \hn \fig{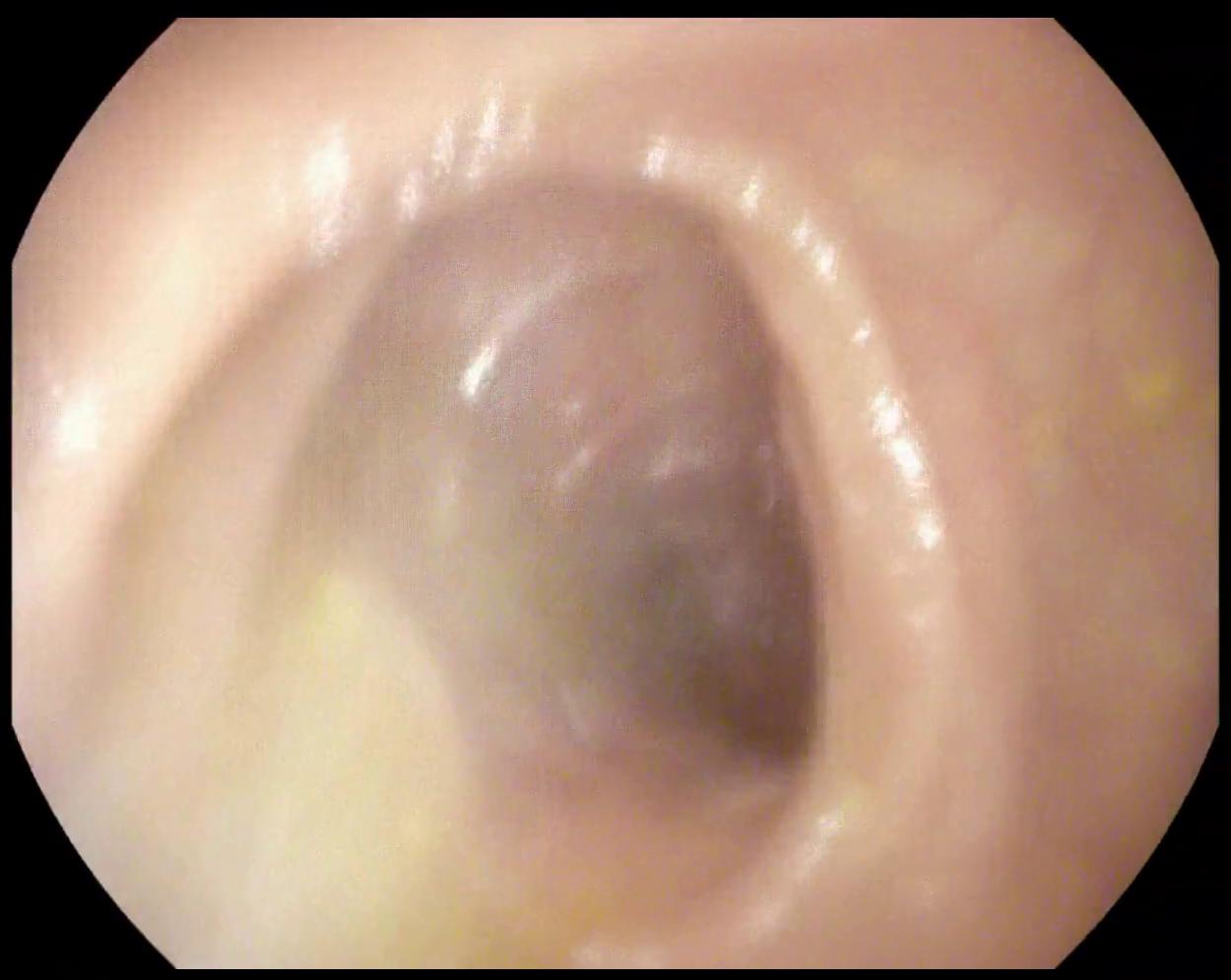}{2.0} & \hn \fig{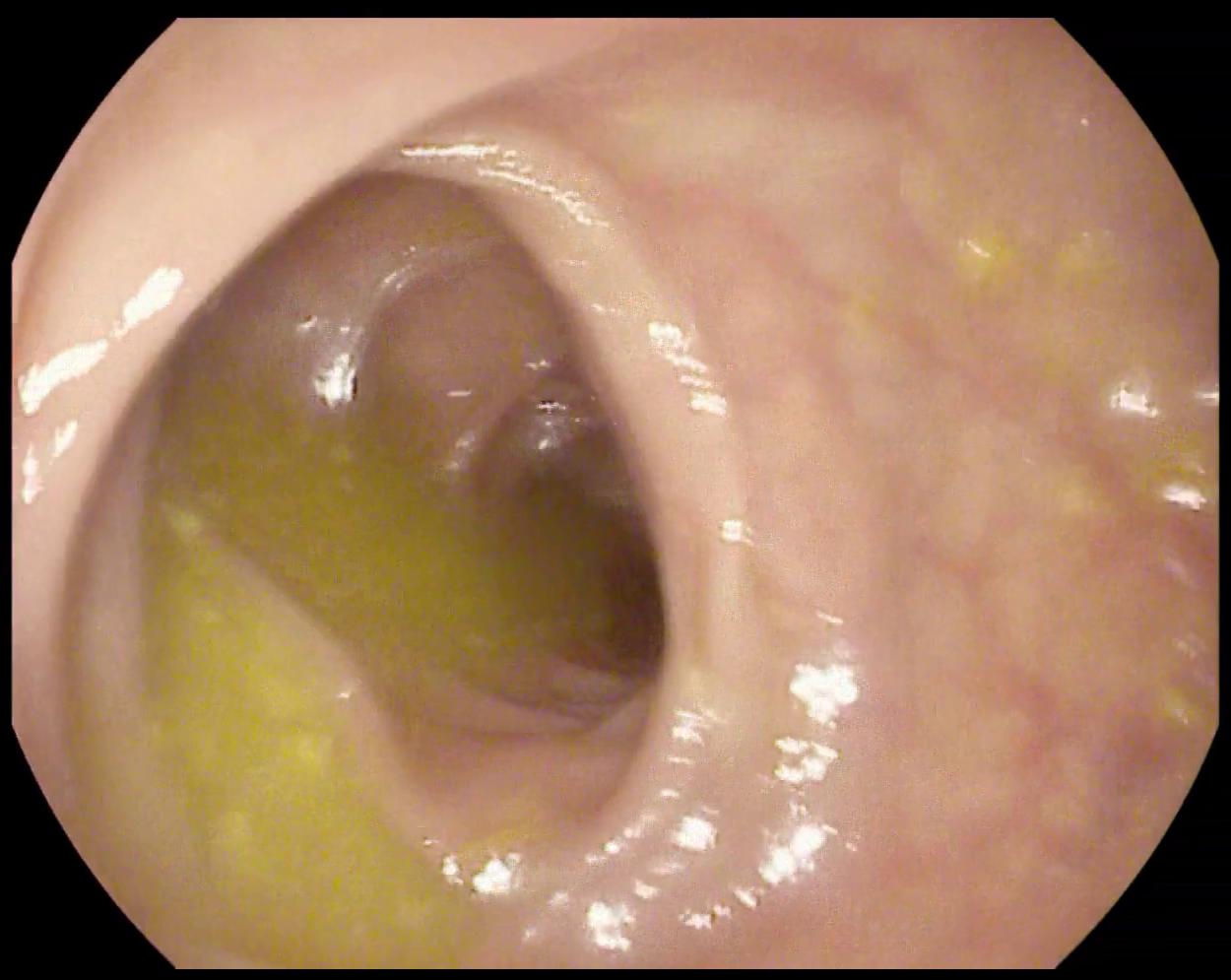}{2.0} & \hn \fig{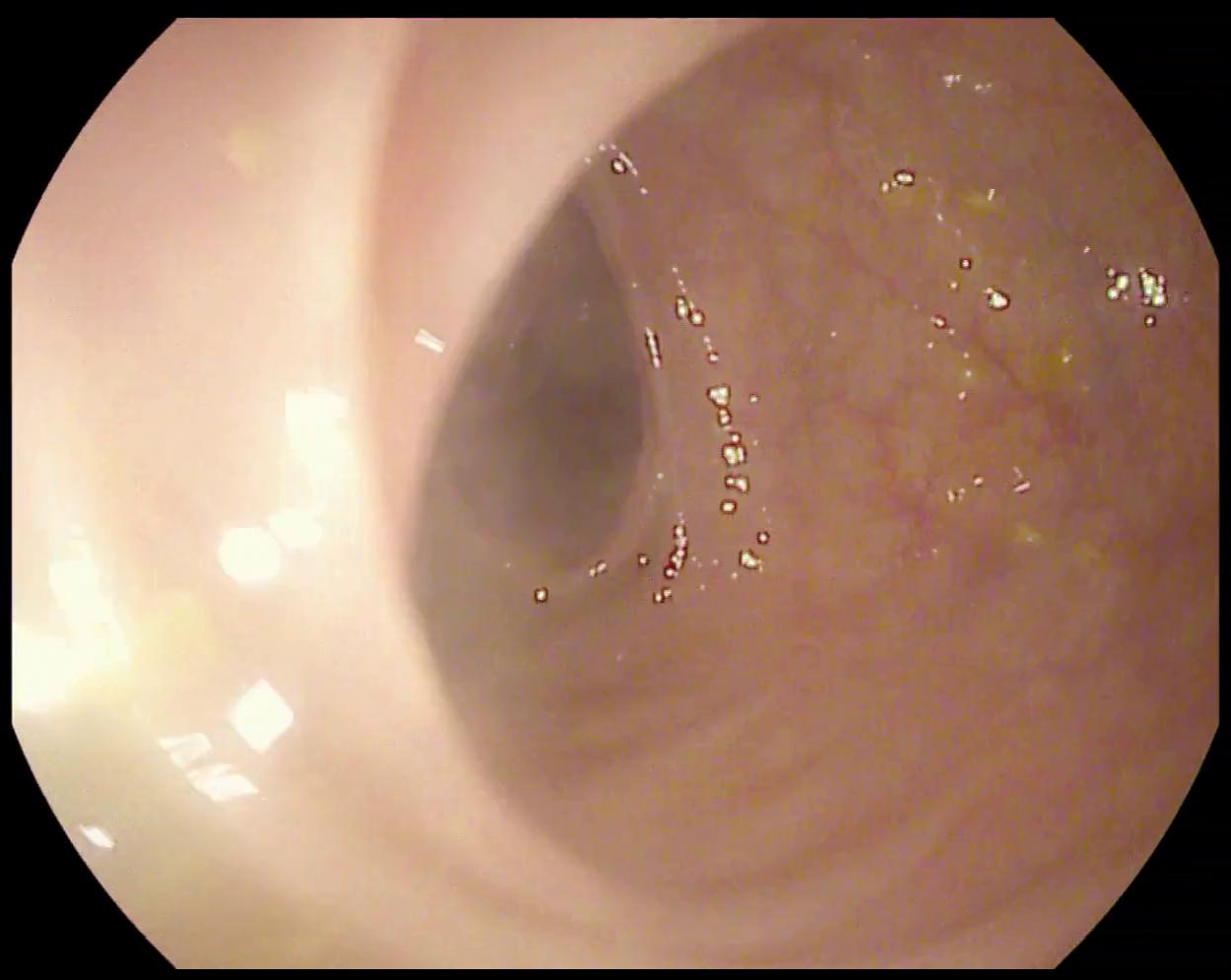}{2.0} \\
  \hnb \rotatebox{90}{\hspace{0.35cm} \scriptsize{0.713}} & \hn \fig{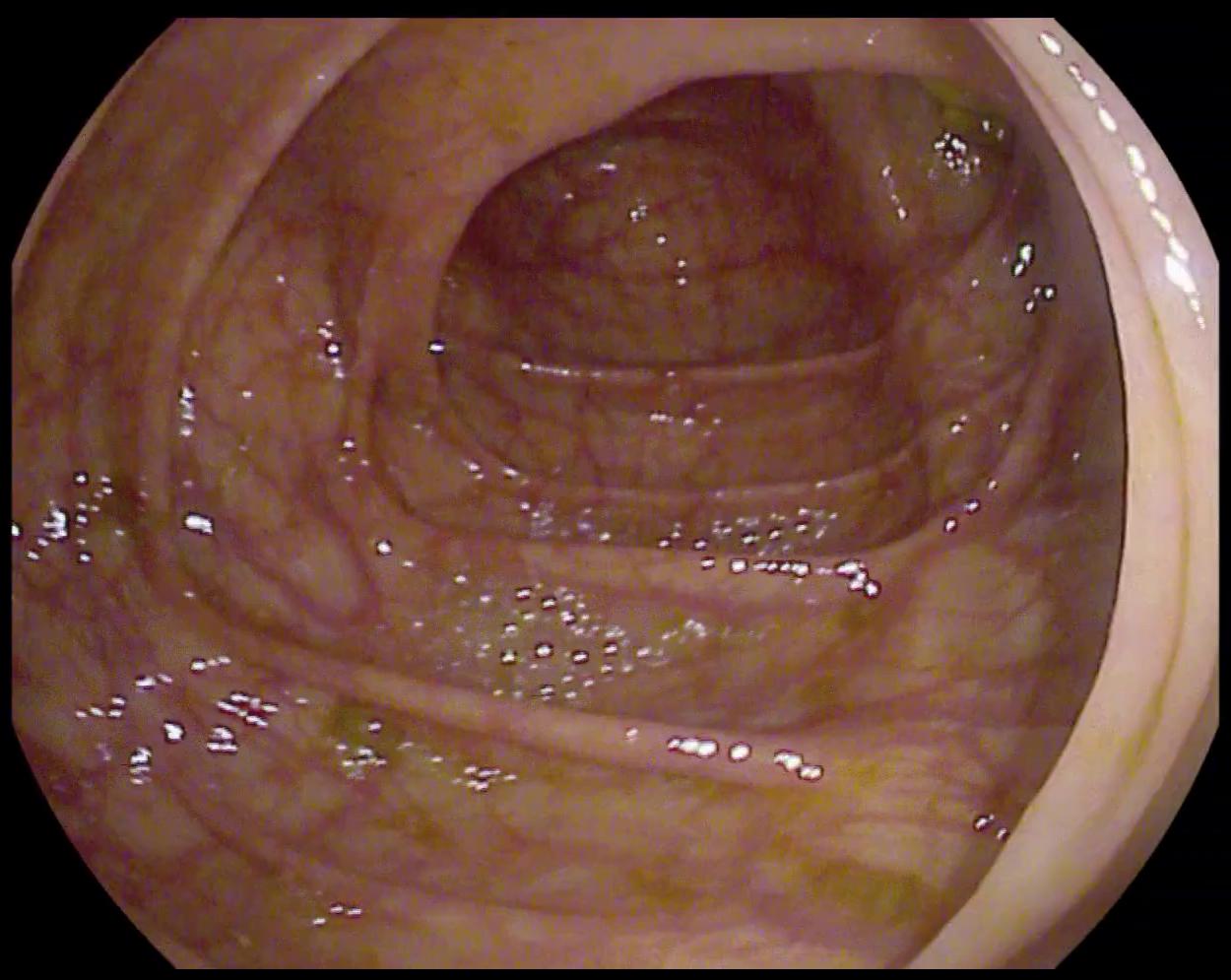}{2.0} & \hn \fig{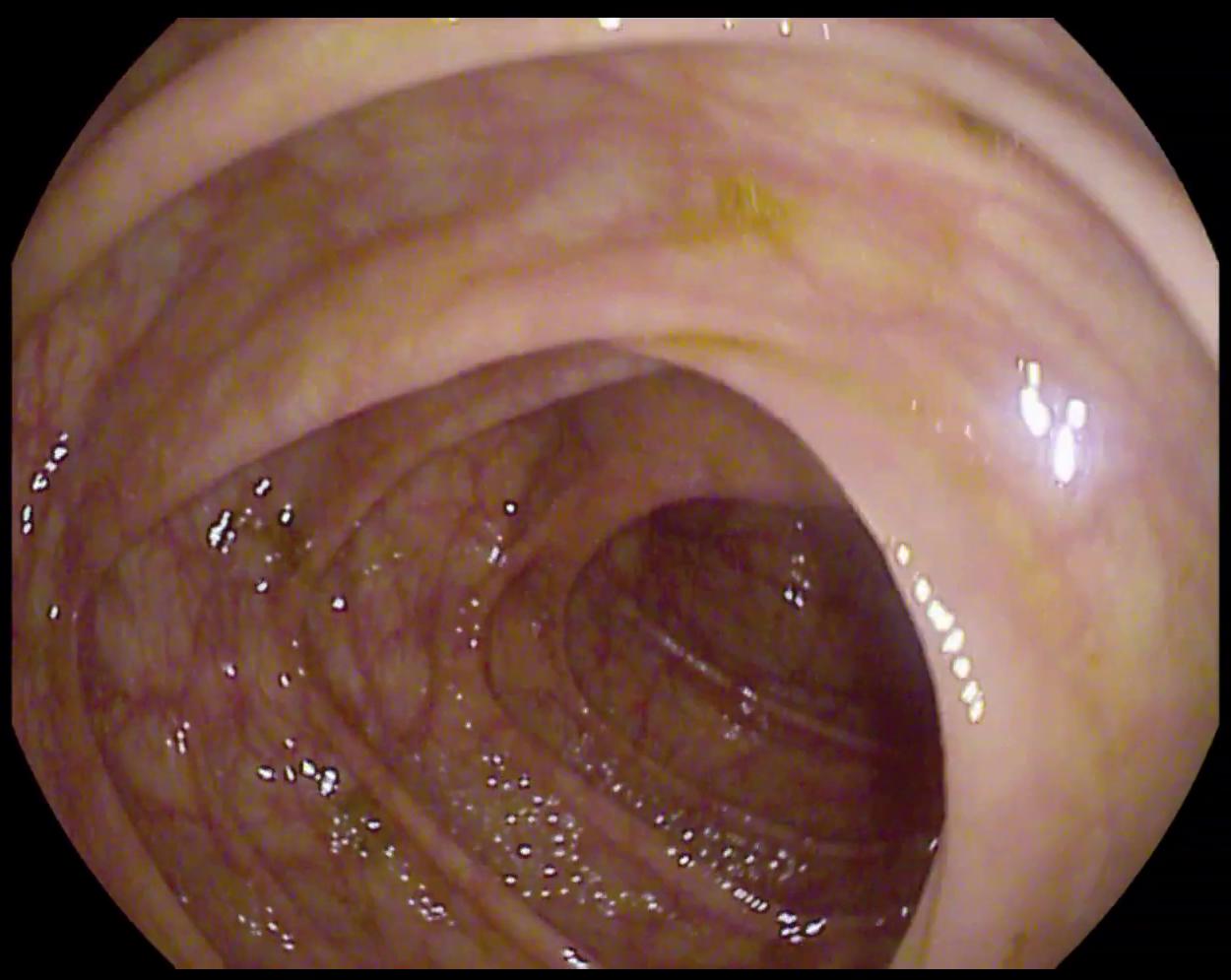}{2.0} & \hn \fig{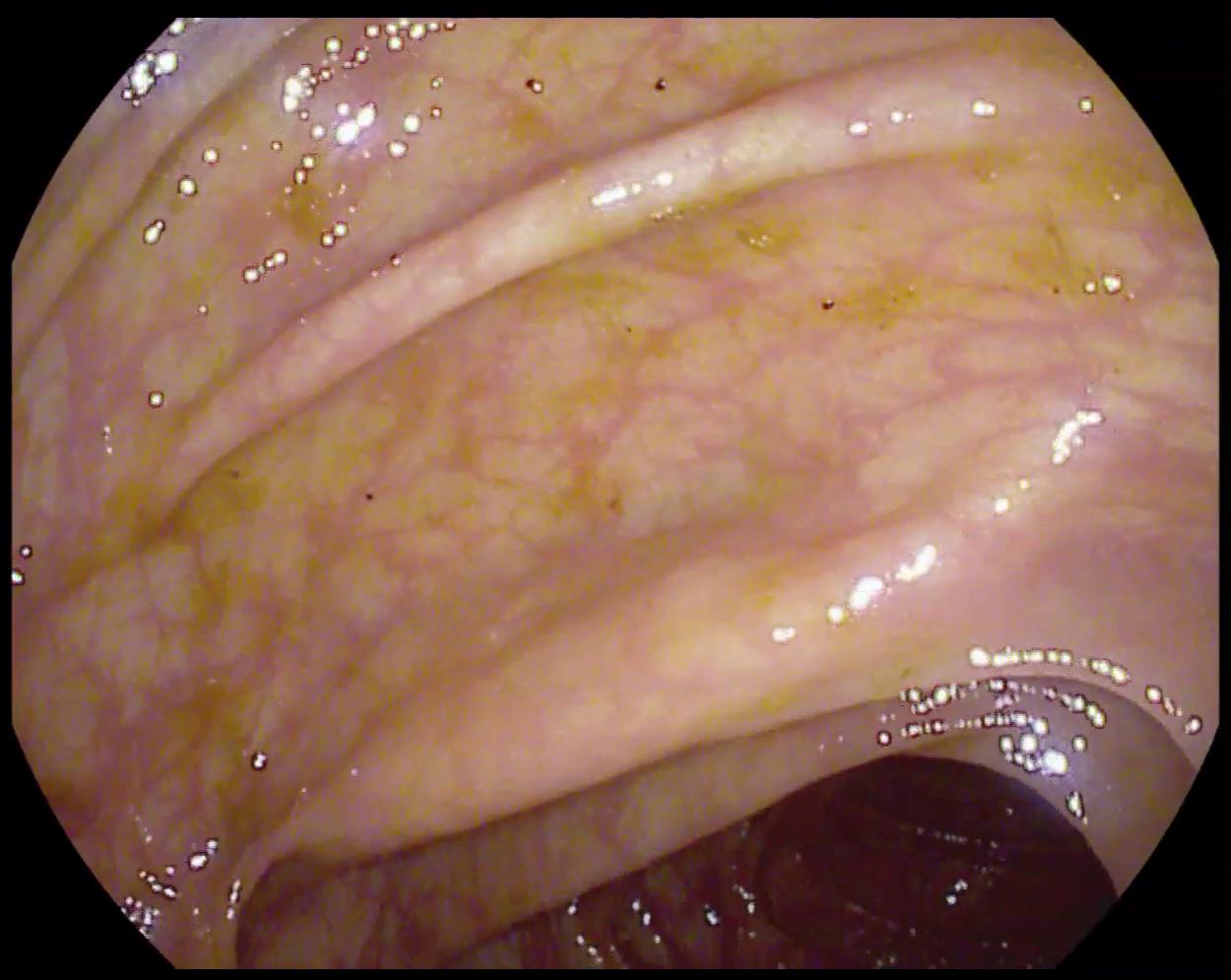}{2.0} & \hn \fig{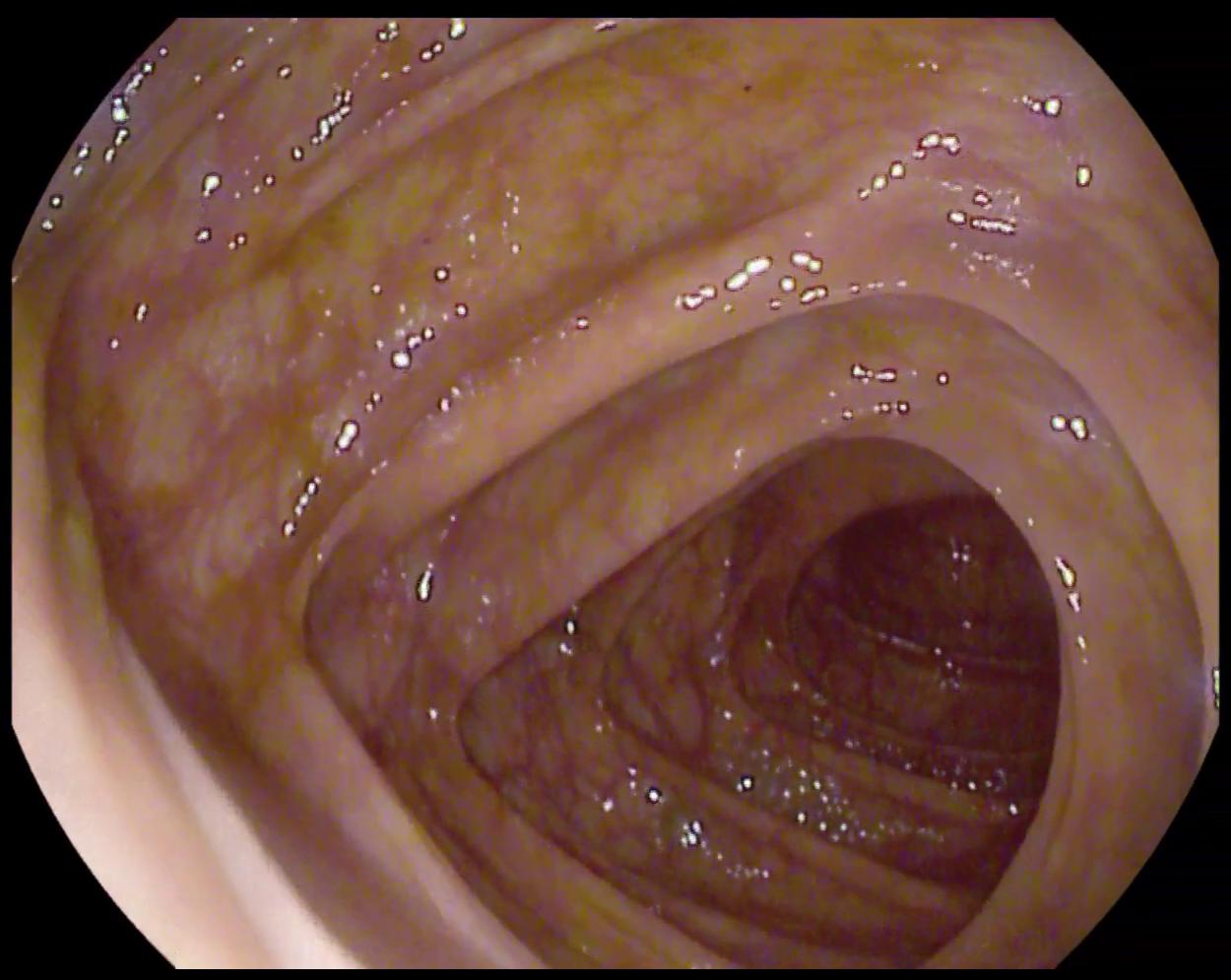}{2.0} \\
  \hnb \rotatebox{90}{\hspace{0.35cm} \scriptsize{0.427}} & \hn \fig{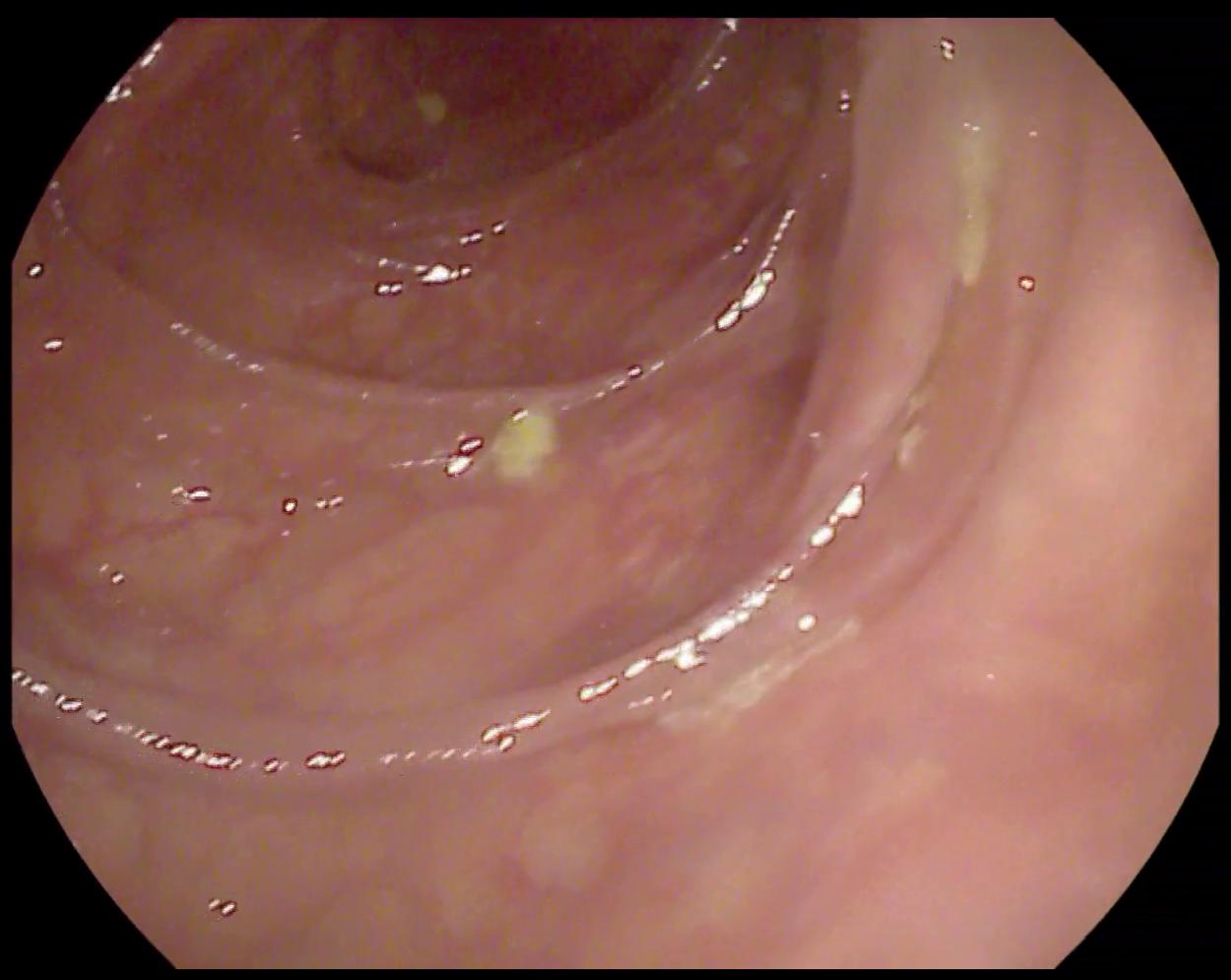}{2.0} & \hn \fig{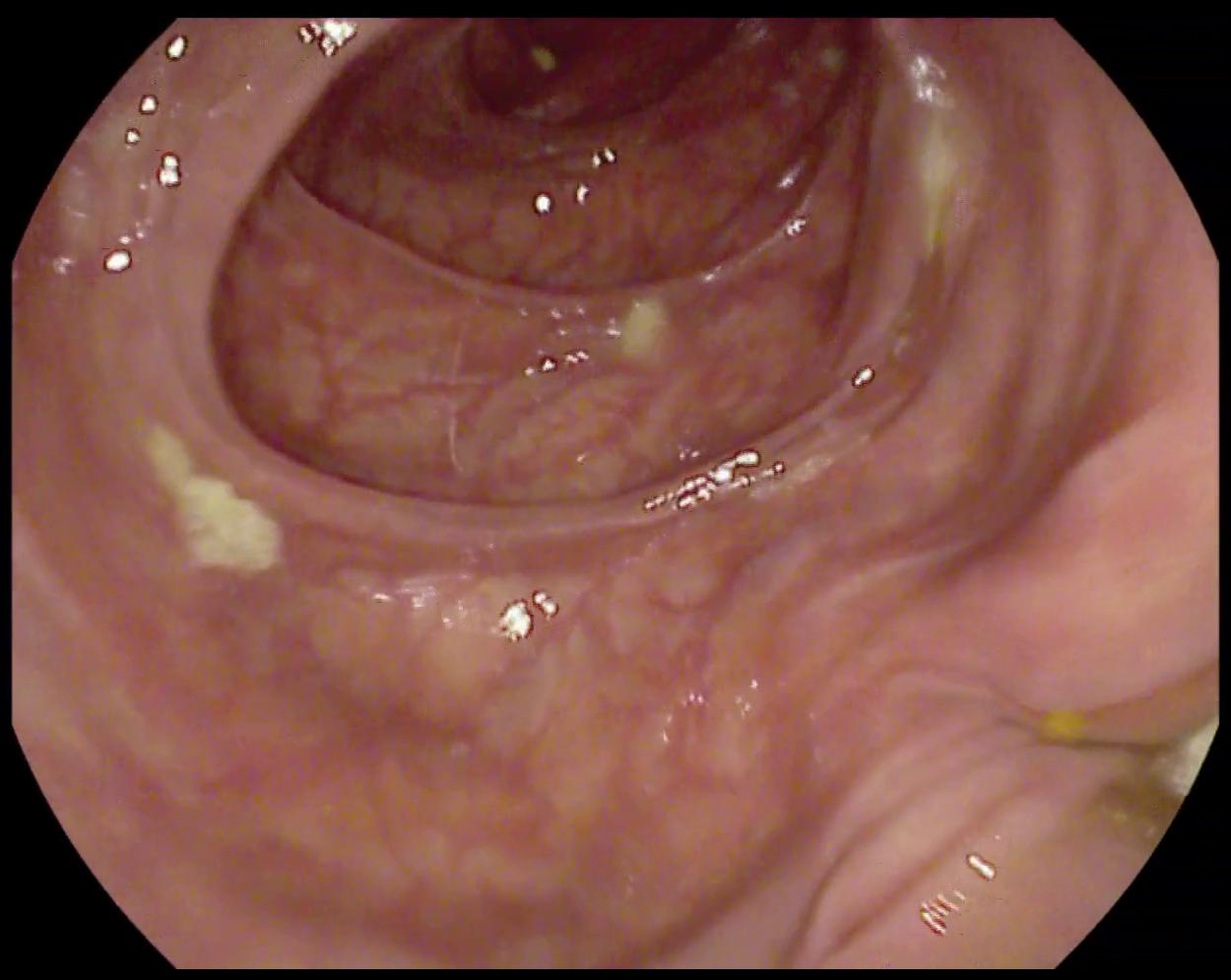}{2.0} & \hn \fig{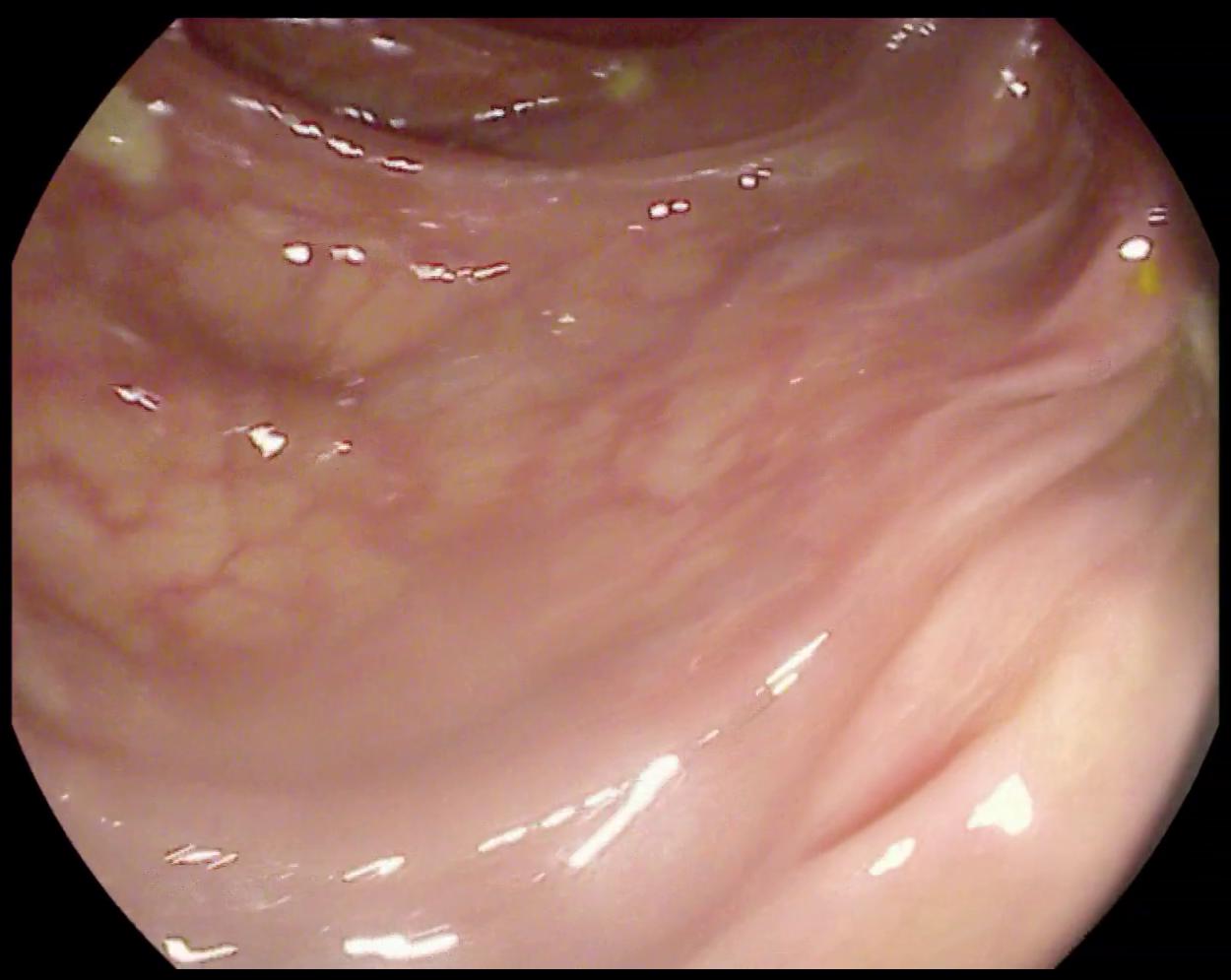}{2.0} & \hn \fig{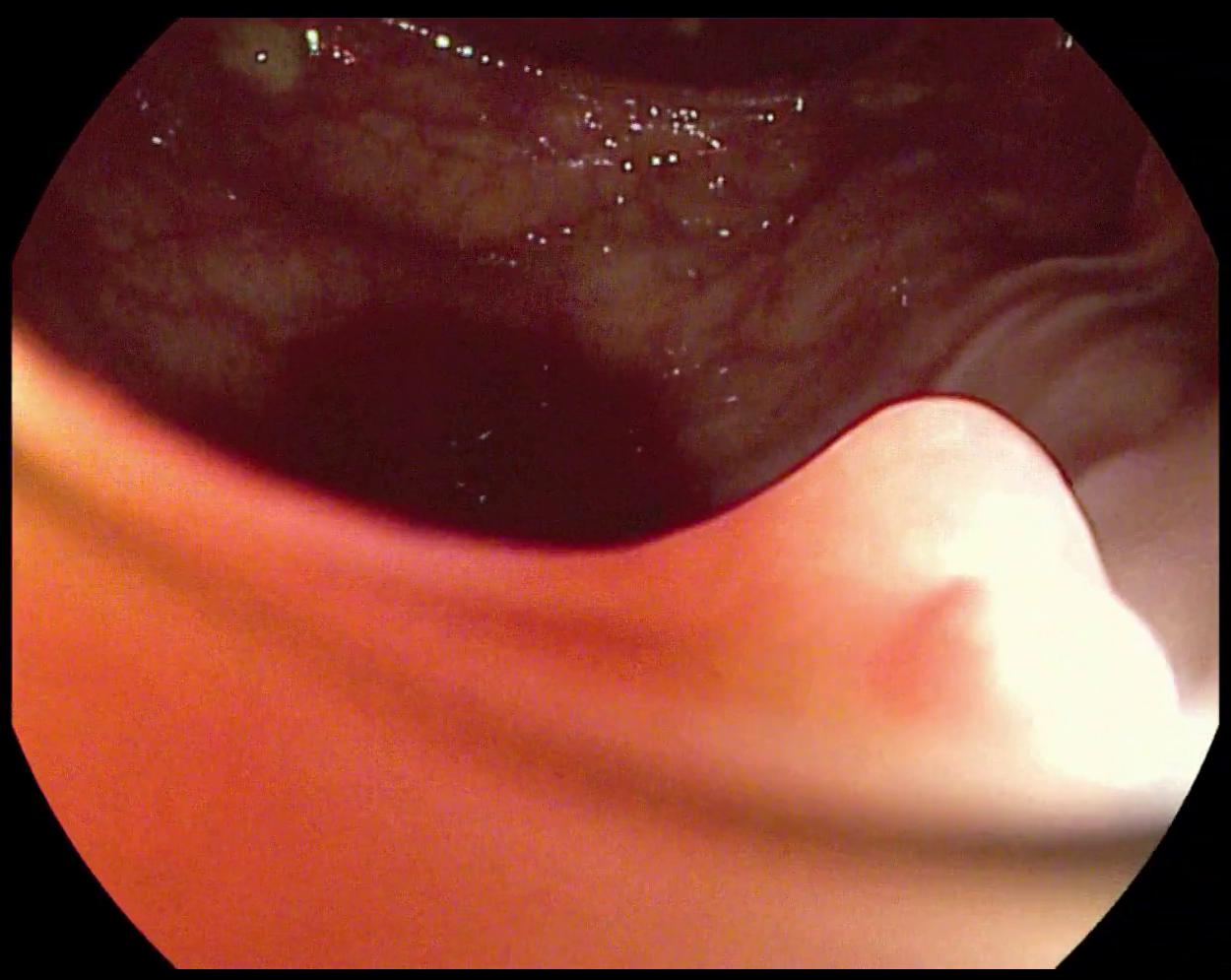}{2.0} \\
  \hnb \rotatebox{90}{\hspace{0.35cm} \scriptsize{0.365}} & \hn \fig{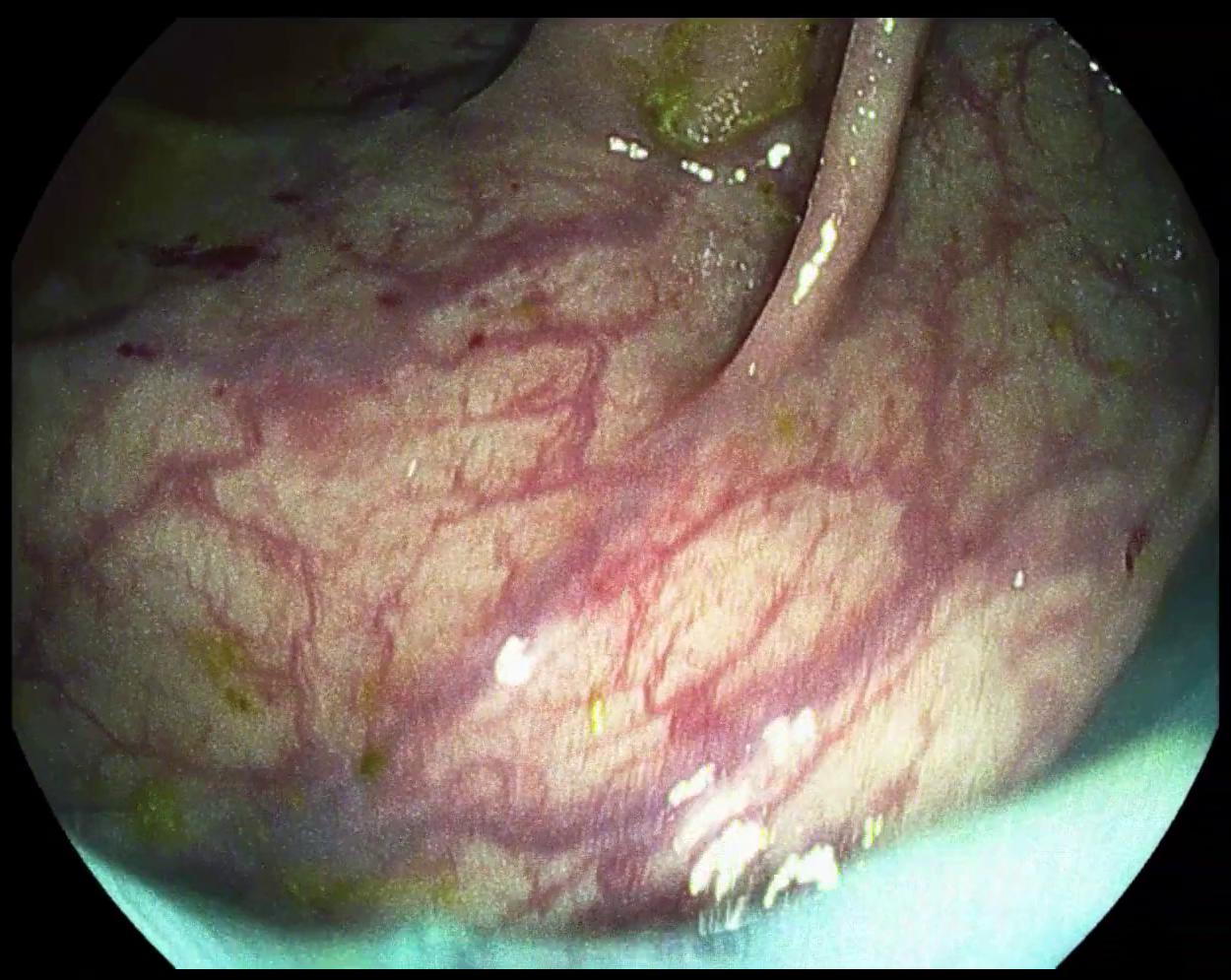}{2.0} & \hn \fig{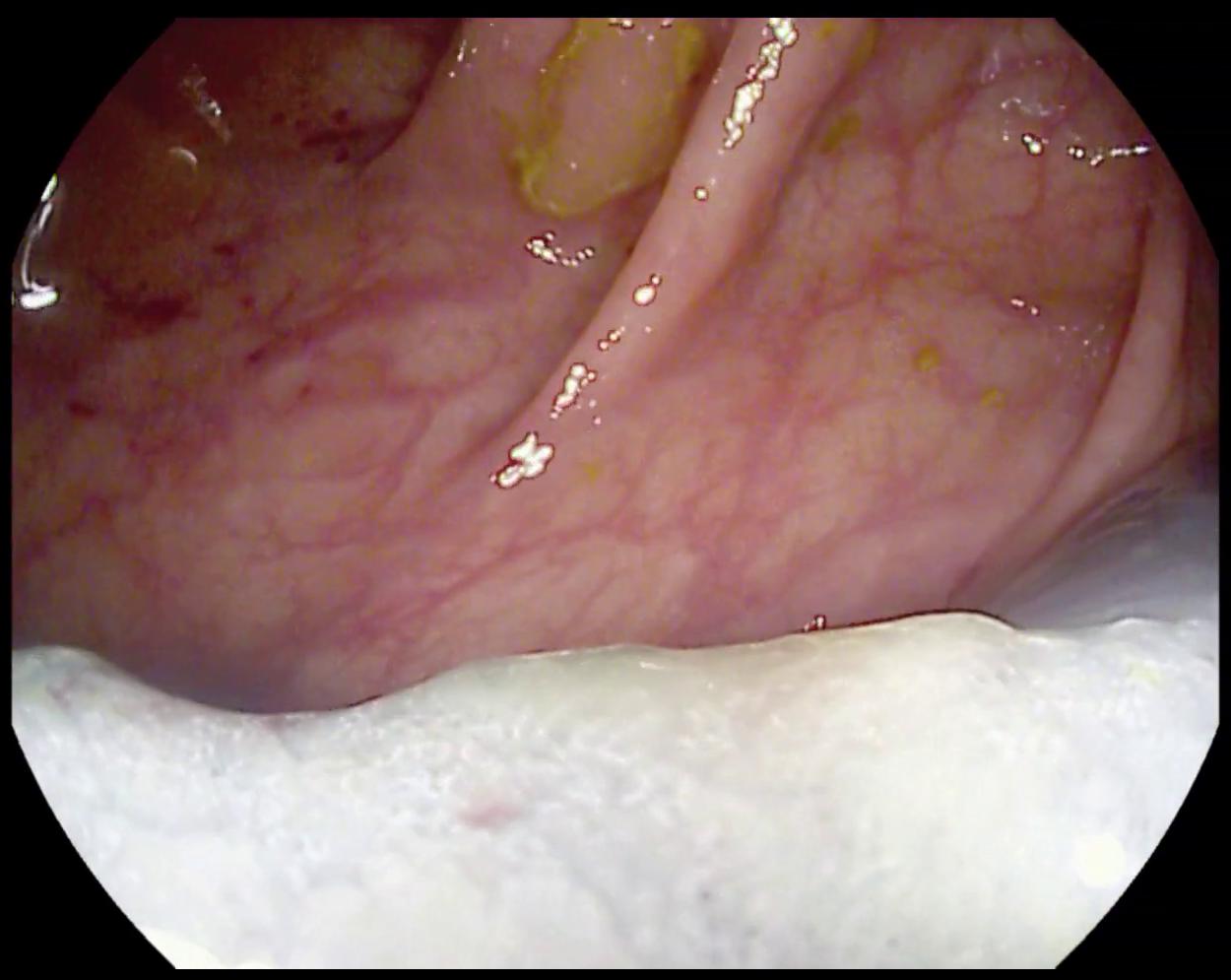}{2.0} & \hn \fig{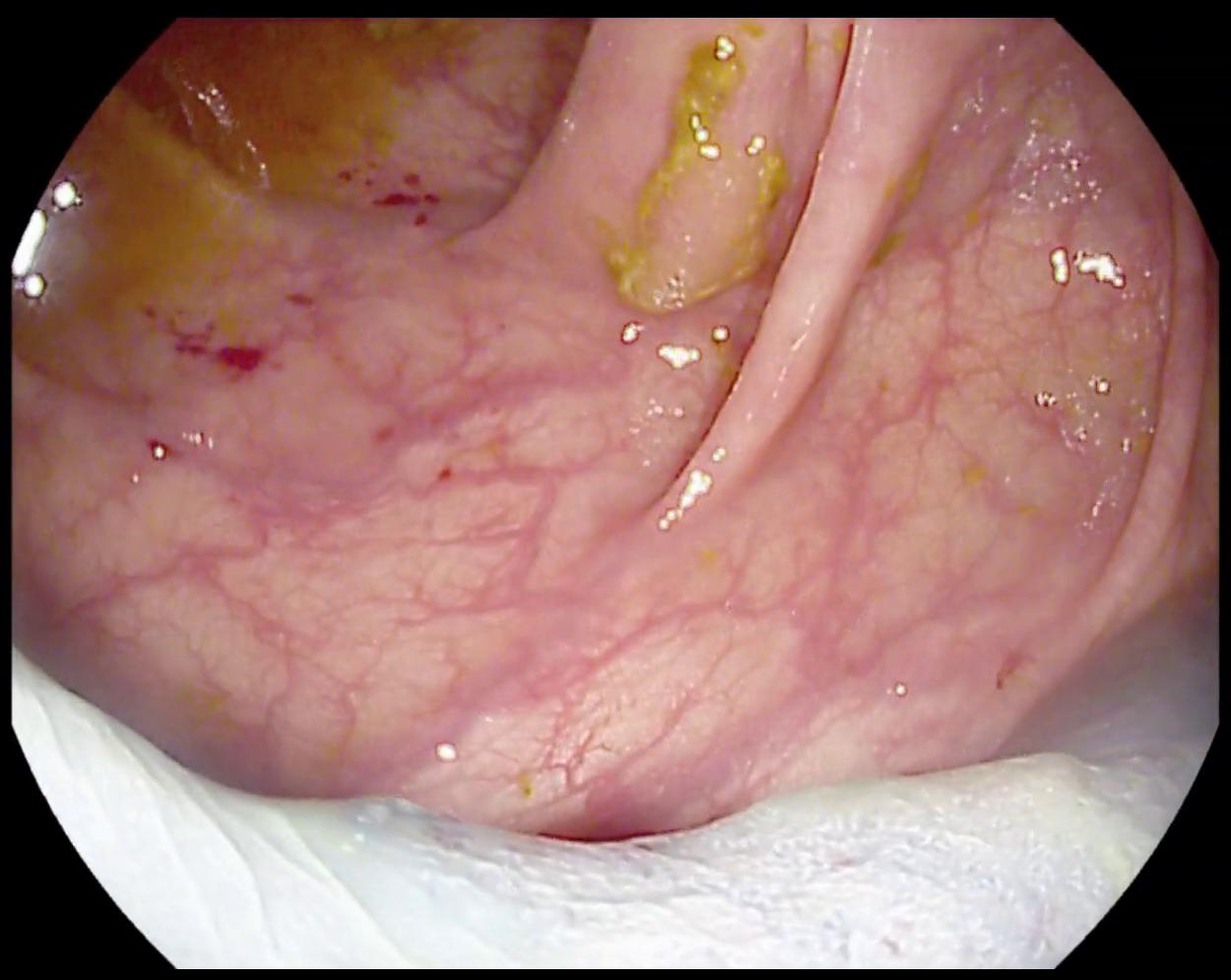}{2.0} & \hn \fig{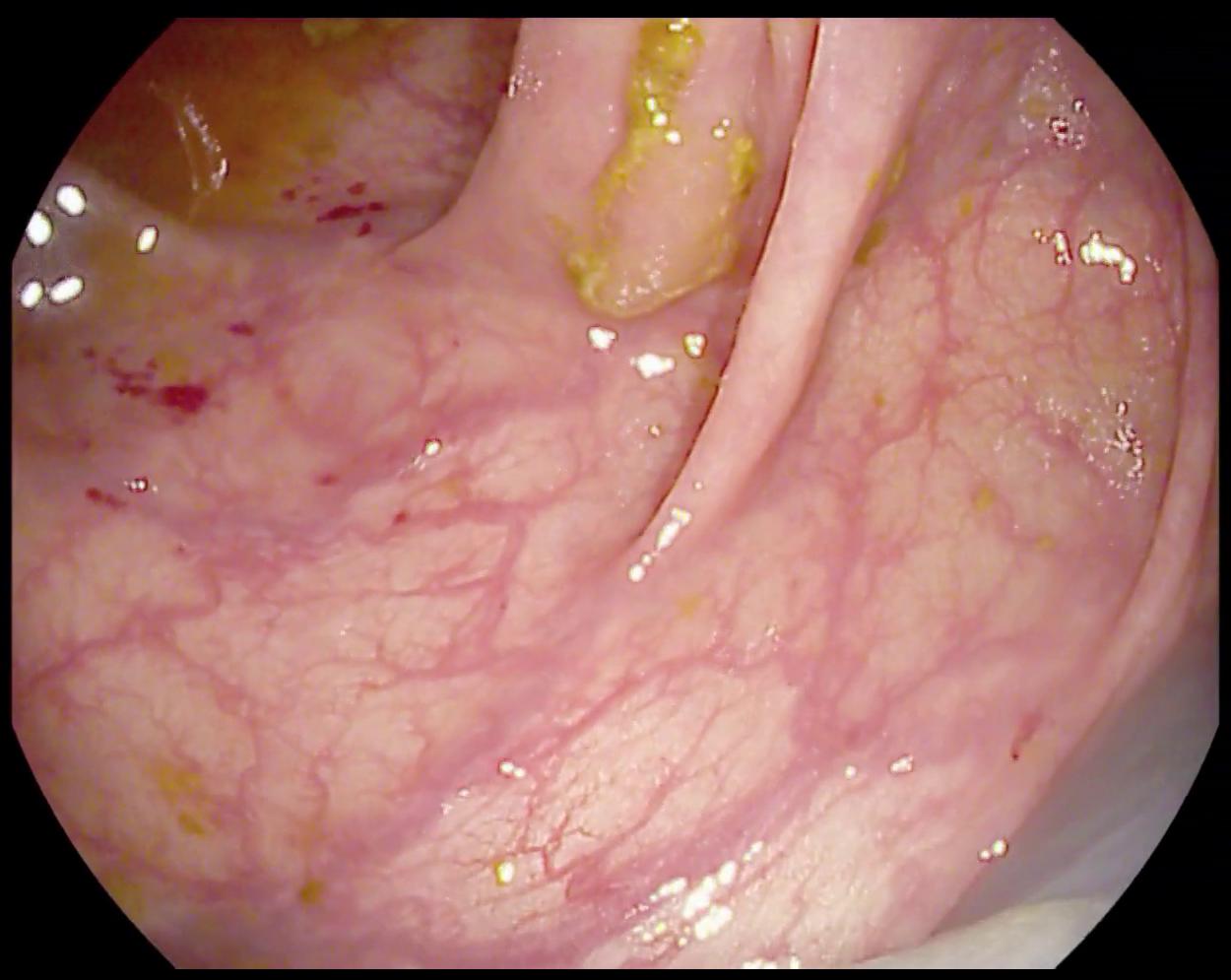}{2.0} \\
  \hnb \rotatebox{90}{\hspace{0.35cm} \scriptsize{0.227}} & \hn \fig{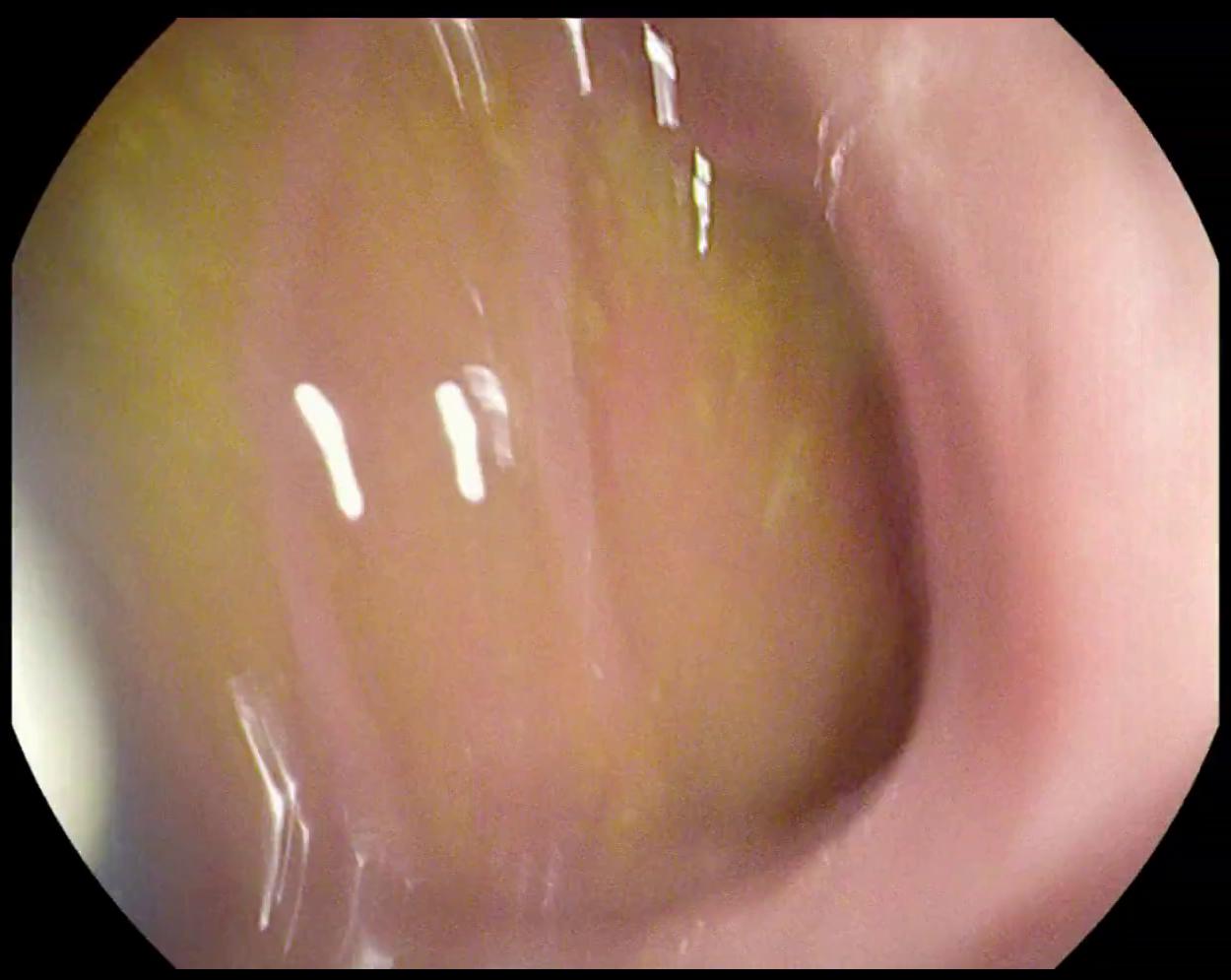}{2.0} & \hn \fig{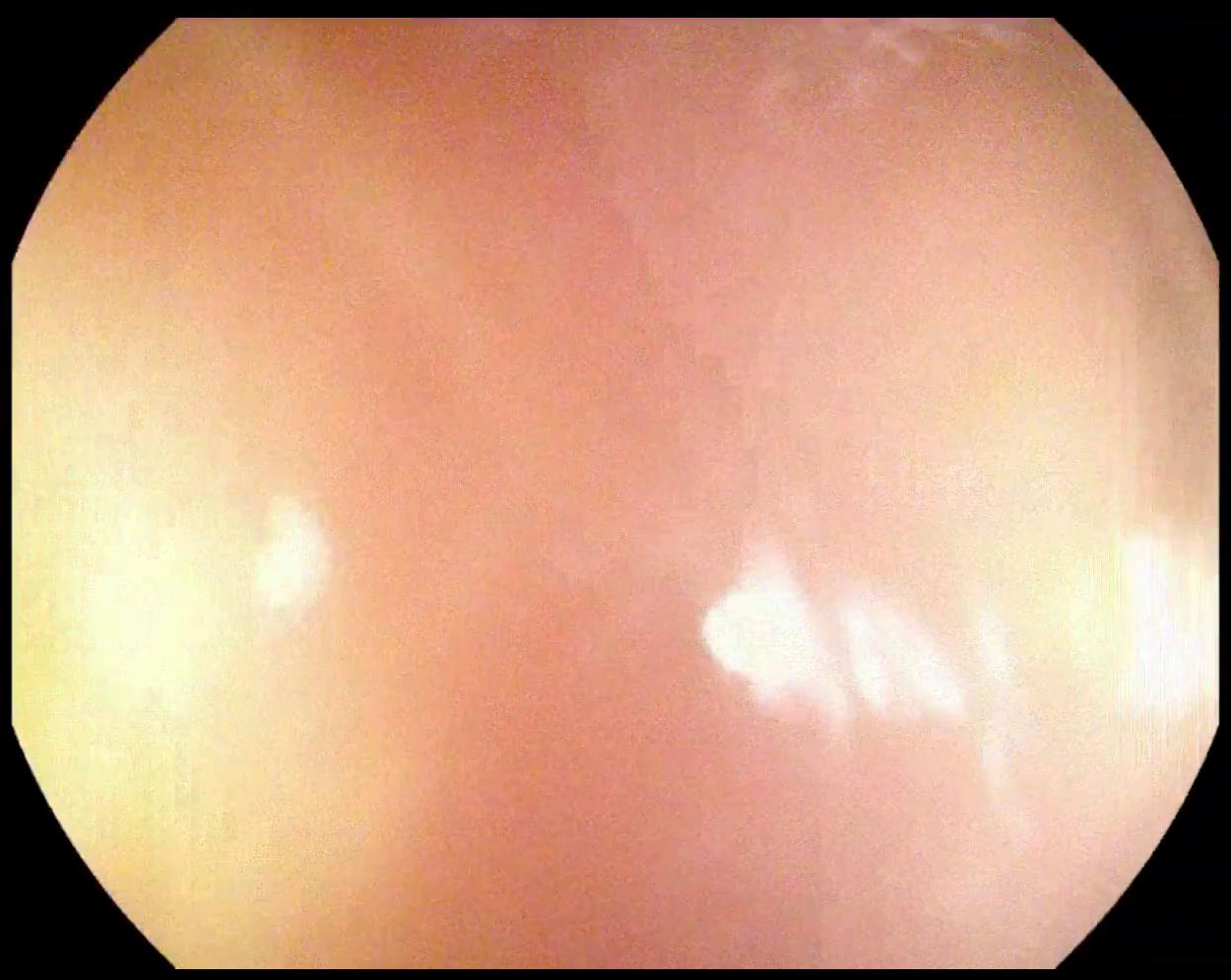}{2.0} & \hn \fig{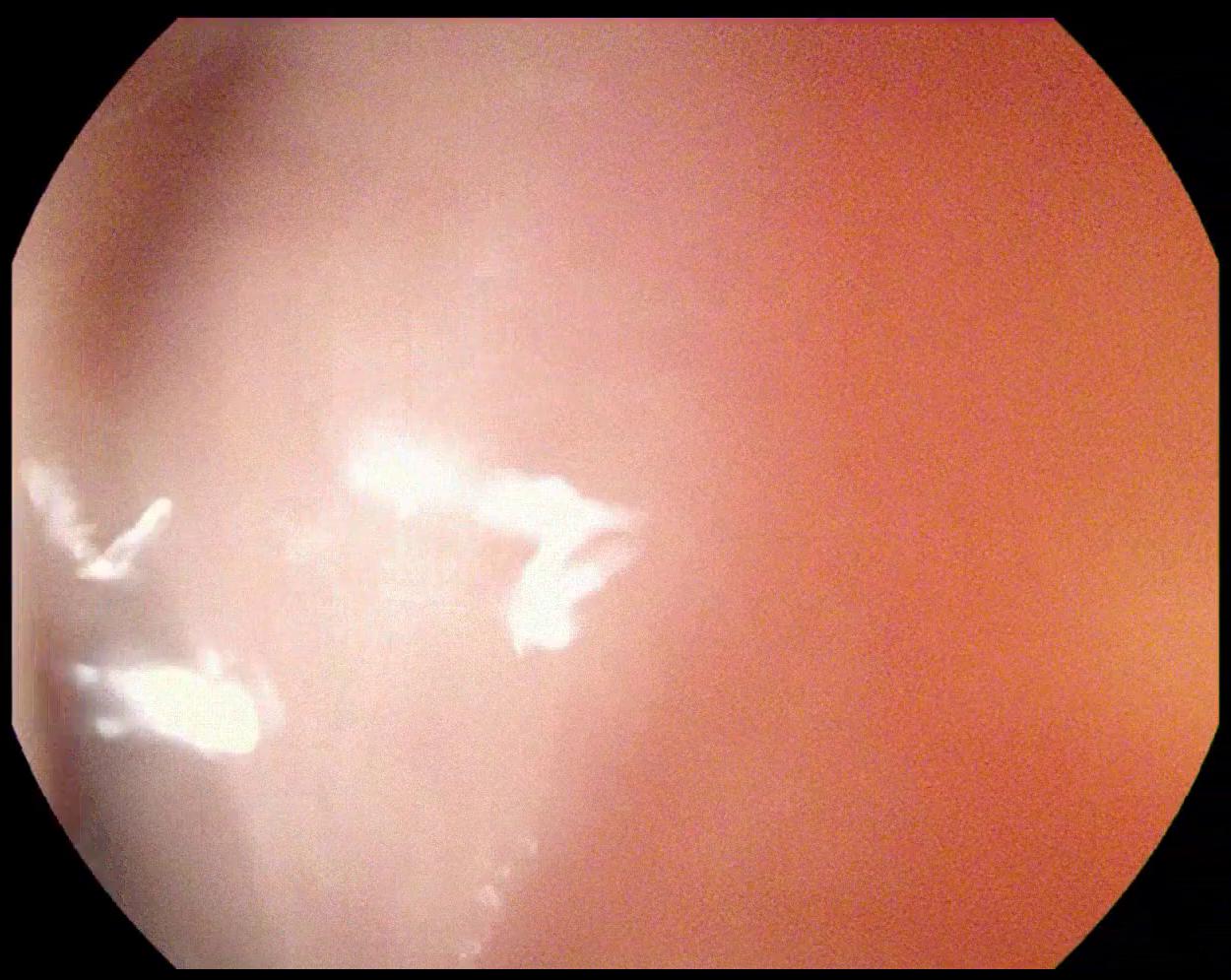}{2.0} & \hn \fig{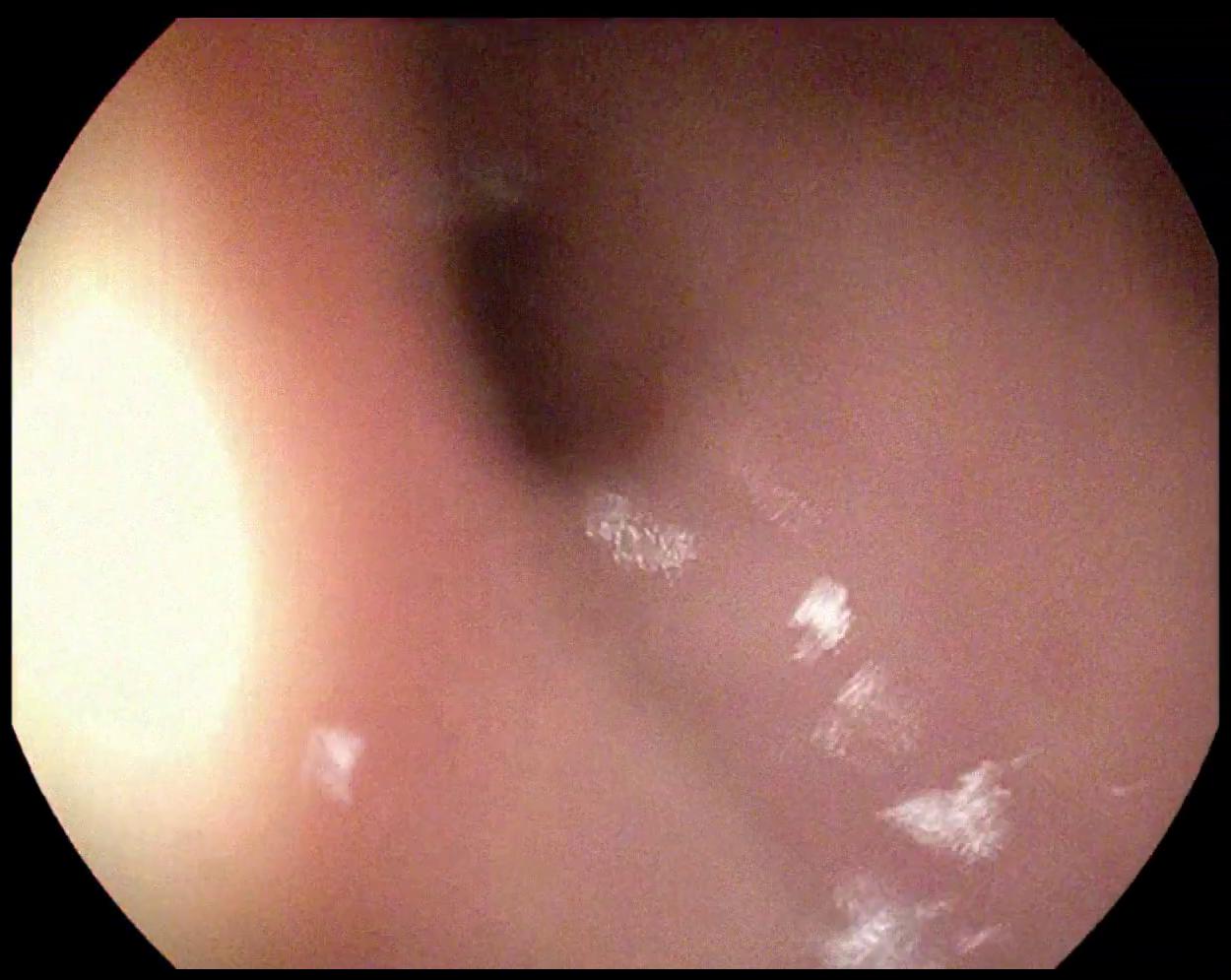}{2.0}
  \end{tabular}
  \caption{Examples of C2D2's score on real sequences; in each case, we show four frames from the sequence, and we report C2D2's coverage score on the left of the row.  The rows are arranged in order of descending coverage score.  See accompanying description in the text.}
  \label{fig:per_segment_qualitative}
\end{figure}

Finally, we turn to qualitative performance on real video segments; representative results are shown in Figure \ref{fig:per_segment_qualitative}, with the rows arranged in order of descending C2D2 coverage score.  The top two rows show examples of high coverage scores; C2D2 reports scores of 0.931 and 0.920.  In each of these cases, it is clear that one can easily see the lumen, and the ``tunnel'' going down the center of the colon.  The second row is interesting, in that the sequence is not as clean as that of the first row: the images are blurry, and there is also fecal material present.  Nevertheless, C2D2 succeeds in reporting a high score.  The third row shows a colon which is mostly covered, but it is clear that the sight line to the lumen is not straight on, and therefore various parts of the colon are missed to some degree (this is particularly noticeable in the third frame); C2D2's score of 0.713 is therefore quite plausible.  The fourth row shows a partially covered colon: the bottom may be seen clearly, but the top is not visible.  C2D2 accordingly gives a score of 0.427.  The fifth row shows a somewhat similar example, except now more parts of the colon are clearly less visible: one cannot see the lumen at all, as compared to the third row, where part of it is somewhat visible.  C2D2 assigns a lower score of 0.365 in this case.  Finally, the sixth row shows an example in which much of the sequence is facing the intestinal wall, with occasional very partial views.  Such a sequence sensibly receives a very low score of 0.227.  In summary, C2D2 succeeds in returning coverage scores on real sequences which pass the eyeball test.

\section{Conclusions}
\label{sec:conclusions}
We have presented C2D2, a new technique for computing coverage of a colonoscopy procedure, and we have demonstrated the accuracy of the technique on a large scale dataset.  To the best of our knowledge, this is the first time a coverage algorithm has been evaluated on such a dataset.  
\revisedremove{
Our results show that C2D2 outperforms human experts by a wide margin on synthetic datasets with ground truth, and highly visually plausible results on real videos (where there is no ground truth).
}
\revisedtag{27}{
Our results show that C2D2 outperforms human experts by a wide margin on synthetic datasets with ground truth, and has a 93.0\% agreement with physicians on real videos.
}
Furthermore, as a building block used in achieving the goal of coverage, we have presented a depth estimation algorithm which is the first unsupervised, calibration-free method to be applied in the colonoscopy domain.  This algorithm has also been shown to attain very promising results.  In the future, we plan to test the efficacy of the coverage algorithm in a live clinical setting.

\section*{Acknowledgments}
The authors would like to thank Gaddi Menahem and Yaron Frid of Orpheus Medical Ltd. for helping in the provision of data.  We would also like to thank Ariel Ben Moshe, Eran Negrin, Smadar Magal, and Ran Bronstein of 3D Systems for their great help in describing the details of their colon simulation system.  Finally, we would like to thank all of our team members and collaborators who worked on this project with us: David Ben Shimol, Nadav Rabani, Chen Barshai, Nia Stoykova, Jesse Lachter, and Ori Segol.

\label{sec:references}

\bibliographystyle{ieeetr}

\end{document}